\documentclass[twocolumn,10pt,journal,twoside]{IEEEtran}
\usepackage[cmex10]{amsmath}
\usepackage{amssymb}
\usepackage{amsfonts}

\usepackage{amsthm}
\usepackage{algorithm}
\usepackage{algpseudocode}
\usepackage{graphicx}
\usepackage{epsfig}
\usepackage{cite}
\usepackage{tensor}
\usepackage[caption=false,font=footnotesize]{subfig}
\usepackage{color}
\usepackage{makecell}
\usepackage{float}
\usepackage{comment}
\usepackage{subfig}
\usepackage[dvipsnames]{xcolor}
\usepackage{algpseudocode}
\usepackage{amsthm,amsmath,amssymb}
\usepackage{mathrsfs}
\usepackage{booktabs}
\usepackage{tabu}
\usepackage{array}
\usepackage{booktabs}
\usepackage{multirow}
\usepackage{blindtext}

\usepackage{hyperref}
\hypersetup{hypertex=true,
	colorlinks=true,
	linkcolor=blue,
	anchorcolor=blue,
	citecolor=blue}

\graphicspath{{figures/}}
\DeclareGraphicsExtensions{.eps}
\theoremstyle{definition}
\newtheorem{remark}{Remark}

\newcommand\T{{\hspace{-0pt}\intercal}}

\DeclareMathOperator{\diag}{diag}

\DeclareMathOperator{\mean}{mean}

\renewcommand{\baselinestretch}{0.98}

\begin{document}

\title{Model Predictive Manipulation of Compliant Objects with Multi-Objective Optimizer and Adversarial Network for Occlusion Compensation}

\author{Jiaming Qi,
        Dongyu Li,
        Yufeng Gao,
        Peng Zhou, and
		David Navarro-Alarcon,~\IEEEmembership{Senior~Member,~IEEE}

\thanks{This work is supported in part by the Research Grants Council (RGC) of Hong Kong under grants 14203917 and 15212721, in part by the Key-Area Research and Development Program of Guangdong Province under Grant 2020B090928001, and in part by the Jiangsu Industrial Technology Research Institute Collaborative Funding Scheme under grant 43-ZG9V.}%
\thanks{J. Qi and Y. Gao are with the Harbin Institute of Technology, Department of Control Science and Engineering, Harbin, Heilongjiang, China. (e-mail: 18B904030@hit.edu.cn,  gaoyf@stu.hit.edu.cn)}%
\thanks{D. Li is with Beihang University, School of Cyber Science and Technology, Beijing, China. (e-mail: dongyuli@buaa.edu.cn)}%
\thanks{P. Zhou and D. Navarro-Alarcon are with The Hong Kong Polytechnic University, Department of Mechanical Engineering, Kowloon, Hong Kong. (e-mail: jeffery.zhou@connect.polyu.hk, dna@ieee.org)}%
}


\markboth{}
{Qi \MakeLowercase{\textit{et al.}}: MPC-Based Manipulation of Compliant Objects with Multi-Objective Optimizer and Occlusion Compensation Network}
\maketitle

\begin{abstract}
	The robotic manipulation of compliant objects is currently one of the most active problems in robotics due to its potential to automate many important applications. 
	Despite the progress achieved by the robotics community in recent years, the 3D shaping of these types of materials remains an open research problem.
	In this paper, we propose a new vision-based controller to automatically regulate the shape of compliant objects with robotic arms.
	Our method uses an efficient online surface/curve fitting algorithm that quantifies the object's geometry with a compact vector of features; This feedback-like vector enables to establish an explicit shape servo-loop.
	To coordinate the motion of the robot with the computed shape features, we propose a receding-time estimator that approximates the system's sensorimotor model while satisfying various performance criteria.
	A deep adversarial network is developed to robustly compensate for visual occlusions in the camera's field of view, which enables to guide the shaping task even with partial observations of the object. 
	Model predictive control is utilized to compute the robot's shaping motions subject to workspace and saturation constraints.
	A detailed experimental study is presented to validate the effectiveness of the proposed control framework.
\end{abstract}

\IEEEpeerreviewmaketitle

\section{Introduction}\label{section1}
\IEEEPARstart{T}{he} manipulation/shaping of deformable bodies by robots is a fundamental problem that has recently attracted the attention of many researchers \cite{zhu2021challenges}; Its complexity has forced researchers to develop new methods in a wide range of fundamental areas that include representation, learning, planning, and control.
From an applied research perspective, this challenging problem has shown great potential in various economically-important tasks such as the assembly of compliant/delicate objects \cite{li2018vision}, surgical/medical robotics \cite{cao2020sewing}, cloth/fabric folding \cite{hu2018three}, etc. 
The manipulation of compliant materials contrast with its rigid-body counterpart in that physical interactions will invariably change the object's shape, which introduces additional degrees-of-freedom to the typically unknown objects, and hence, complicates the manipulation task.
While great progress has been achieved in recent years, the development of these types of embodied manipulation capabilities is still largely considered an open research problem in robotics and control.

There are various technical issues that hamper the implementation of these tasks in real-world unstructured environments, which largely differ from implementations in ideal simulation environments (a trend followed by many recent works).
Here, we argue that to effectively servo-control the shape of deformable materials in the field, a sensor-based controller must possess the following features:
1) Efficient compression of the object's high-dimensional shape;
2) Occlusion-tolerant estimation of the objects geometry;
3) Adaptive prediction of the differential shape-motion model;
Our aim in this paper is precisely to develop a new shape controller endowed with all the above-mentioned features. 
The proposed method is formulated under the model predictive control (MPC) framework that enables to compute shaping actions that satisfy multiple performance criteria (a key property for real engineering applications).

\subsection{Related Work}
Many researchers have previously studied this challenging problem (we refer the reader to \cite{Sanchez2018,yin2021modeling} for comprehensive reviews).
To servo-control the object's non-rigid shape, it is essential to design a low-dimensional feature representation that can capture the key geometric properties of the object.
Several representation methods have been proposed before, e.g., geometric features based on points, angles, curvatures, etc  \cite{navarro2013model,navarro2014visual}; However, due to their hard-coded nature, these methods can only be used to represent a single shaping action. 
Other geometric features computed from contours and centerlines \cite{wang2018adaptive,navarro2017fourier,qi2021contour} 
can represent soft object deformation in a more general way.
Various data-driven approaches have also been proposed to represent shapes, e.g., using fast point feature histograms  
\cite{hu20193}, bottleneck layers \cite{qi2021towards,9410363}, principal component analysis \cite{zhu2021vision}, etc.
However, there is no widely accepted approach to compute efficient/compact feature representations for 3D shape; This is still an open research problem.

Occlusions of a camera's field of view pose many complications to the implementation of visual servoing controllers, as the computation of (standard) feedback features requires complete visual observations of the object at all times.
Many methods have been developed to tackle this critical issue, e.g. \cite{cazy2015visual} used the estimated interaction matrix to handle information loss in visual servoing, yet, this method requires a calibrated visual-motor model.
Coherent point drift was utilized in \cite{tang2018framework} to register the topological structure from previous sequences and to predict occlusions; 
However, this method is sensitive to the initial point set, further affects the registration result.
A structure preserved registration method was presented in \cite{tang2018track} to track occluded objects; This approach has good accuracy and robustness to noise, yet, its efficiency decreases with the number of points.
To efficiently implement vision-based strategies in the field, it is essential to develop algorithms that can robustly guide the servoing task, even in the presence of occlusions.


To visually guide the manipulation task, control methods must have some form of model that (at least approximately) describes that how the input robot motions produce output shape changes \cite{navarro2020lyapunov}; In the visual servoing community, such differential relation is typically captured by the so-called interaction (Jacobian) matrix \cite{chaumette2006visual}.
Many methods have been proposed to address this issue, e.g. the
Broyden update rule \cite{hosoda1994versatile} is a classical algorithm to iteratively estimate this transformation matrix \cite{navarro2013model,alambeigi2018autonomous,lagneau2020active}. 
Although these types of algorithms do not require knowledge of the model's structure, its estimation properties are only valid locally.
Other approaches with global estimation properties include algorithms based on (deep) artificial neural networks \cite{li2018vision}, optimization-based algorithms \cite{lagneau2020automatic}, adaptive estimators \cite{han2021visual}, etc.
However, the majority of existing methods estimate this model based on a single performance criterion (typically, a first-order Jacobian-like relation), which limits the types of dynamic responses that the robot can achieve during a task.
A multi-objective model estimation is particularly important in the manipulation of compliant materials, as their mechanical properties are rarely known in practice, thus, making hard to meet various performance requirements.


To compute the active shaping motions, most control methods only formulate the problem in terms of the final target shape and do not typically consider the system's physical constraints. 
MPC represents a feasible solution to these issues, as it performs the control tasks by optimizing cost functions over a finite time-horizon rather than finding the exact analytical solution \cite{hajiloo2015robust};
This allows MPC to compute controls that guide the task while satisfying a set of constraints, e.g., control saturation, workspace bounds, etc.
Despite its valuable and flexible properties, MPC has not been sufficiently studied in the context of shape control of deformable objects.

\subsection{Our Contribution}
To solve the above-mentioned issues, in this paper we propose a new control framework to manipulate purely-elastic objects into desired configurations. 
The original features of our new methodology include:
1) A parametric shape descriptor to efficiently characterize 3D deformations based on online curve/surface fitting;
2) A robust shape prediction network based on adversarial neural networks to compensate visual occlusions;
3) An optimization-based estimator to approximate the deformation Jacobian matrix and satisfy various performance constraints;
4) An MPC-based motion controller to guide the shaping motions while simultaneously solving workspace and saturation constraints.

To the best of the authors' knowledge, this is the first time that a shape servoing controller is developed with all the functions proposed in this paper.
To validate the effectiveness of our new methodology, we report a detail experimental study with a robotic platform manipulating various types of compliant objects.




\begin{figure}[ht]
	\centering
	\includegraphics[scale=0.30]{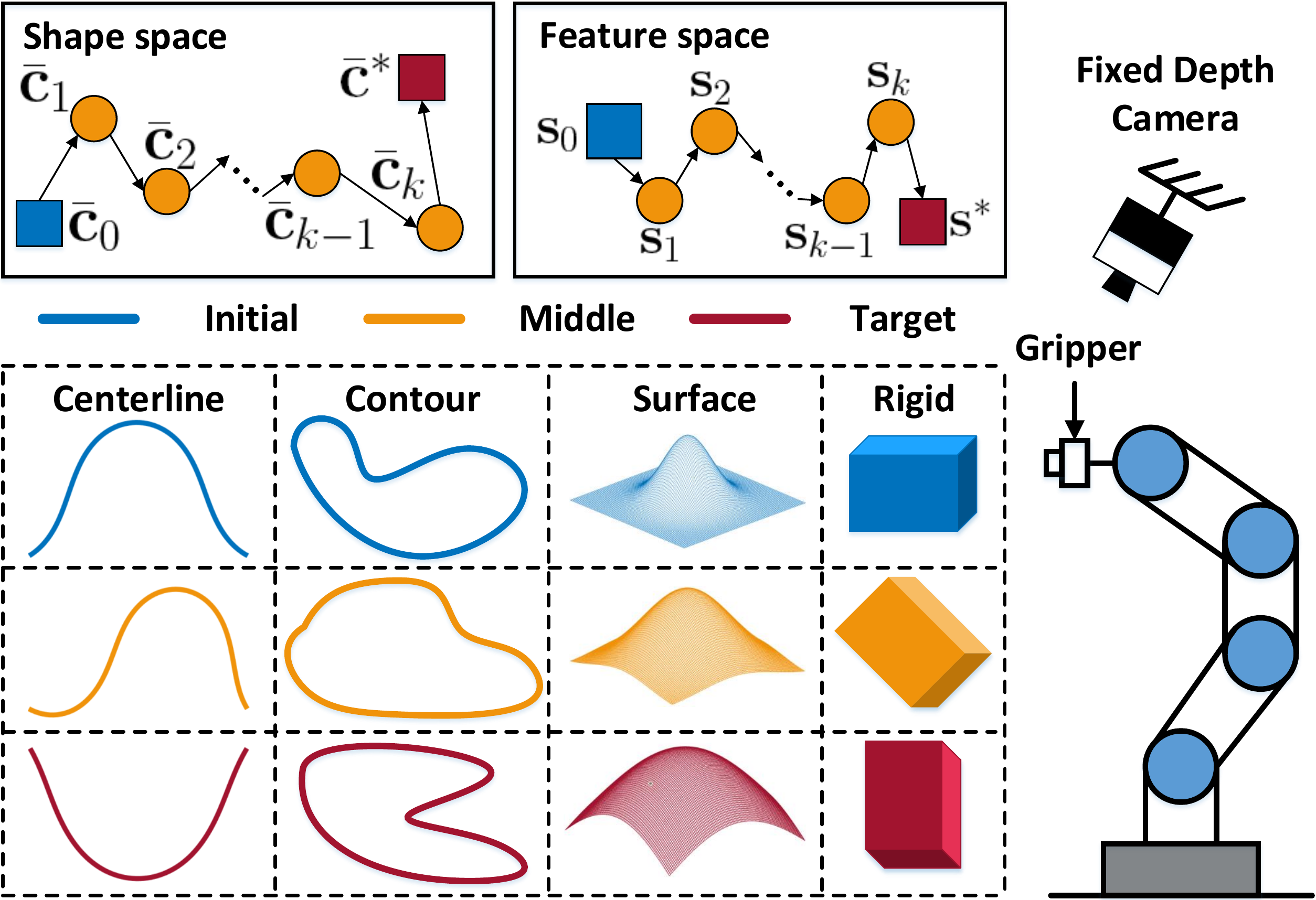}
	\vspace{-0.3cm}
	\caption{
	Schematic diagram of the manipulations of elastic objects, including deformation (centerline, contour, and surface) and positioning tasks of rigid objects.
	The framework aims to command the robot to manipulate elastic objects into the target configurations.}
	\label{fig29}
\end{figure}

\vspace{-0.8cm}

\begin{figure}[ht]
	\centering
	\includegraphics[scale=0.16]{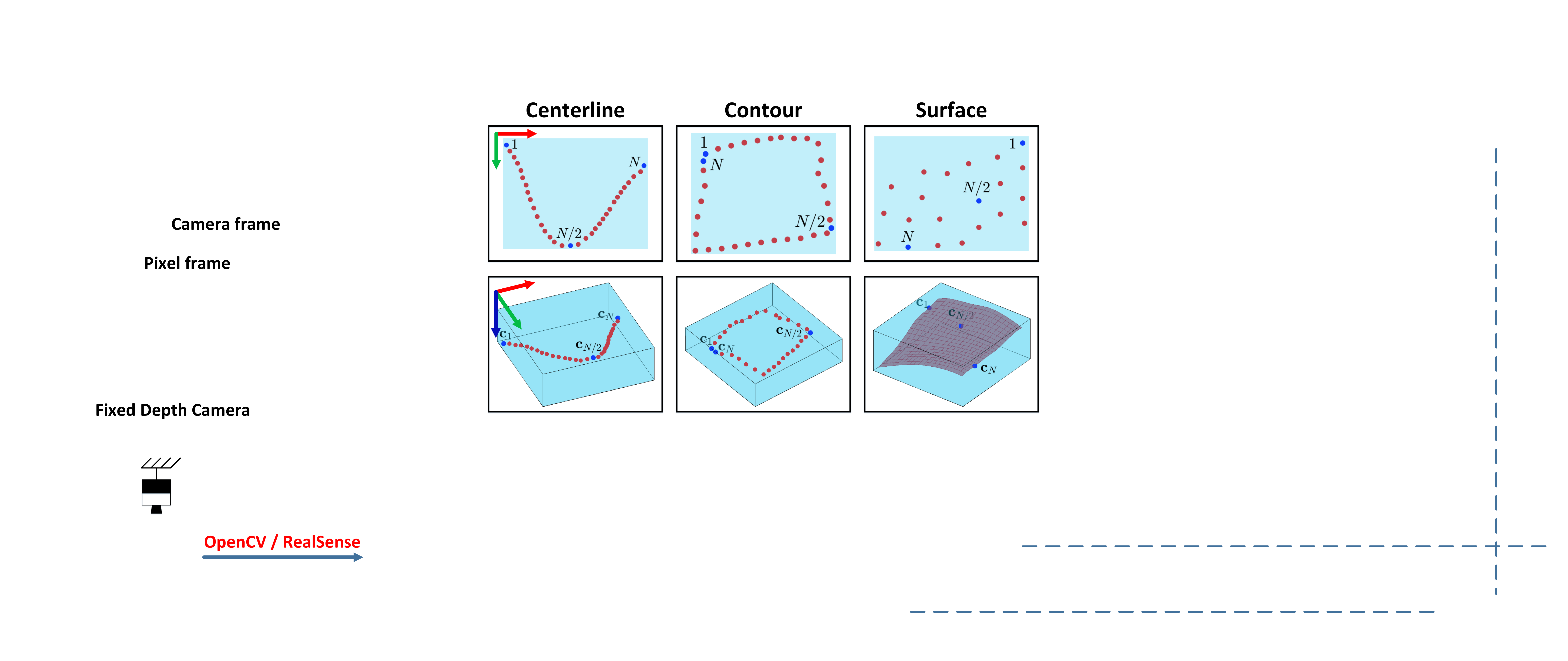}
	\vspace{-0.3cm}
	\caption{
	Various shape configurations, including centerline, contour and surface.
	The first row shows the pixel coordinates for each shape.
	The second row shows the Cartesian coordinates related to the corresponding pixel ones through $OpenCV/RealSense$.
	The generated shapes are ordered, fixed-sampled and equidistant.
	The rigid object adopts surface representation.
	}
	\label{fig30}
	
\end{figure}

\vspace{-0.4cm}
\section{Problem Formulation}\label{section2}
\emph{Notation:} In this paper, we use the following frequently-used notation:
Bold small letters, e.g., $\mathbf{v}$, denote column vectors, while bold capital letters, e.g., $\mathbf{M}$, denote matrices.
Time evolving variables are denoted as $\mathbf{x}_k$, for $k$ as the discrete time instant.
The $n\times m$ matrix of ones is denoted by $\mathbf{I}_{n \times m}$ and the identity matrix as $\mathbf{E}_n$.
$\mathbf{L}_n$ represents the low triangle matrix of $\mathbf{I}_{n \times n}$, and $\otimes$ represents the Kronecker product.


The schematic diagram of the proposed shape servoing framework is conceptually illustrated in Fig. \ref{fig29}.
A depth camera with eye-to-hand configuration observes the shapes of elastic objects that are manipulated by the robot, see Fig. \ref{fig30}.
We denote the 3D measurement points captured by the vision system as:
\begin{equation}
\label{eq1}
\begin{array}{*{20}{c}}
{\bar{\mathbf{c}} = {{\left[ {\mathbf{c}_1^\T, \ldots ,\mathbf{c}_N^\T} \right]}^\T} \in {\mathbb{R}^{3N}}},&
{{\mathbf{c}_i} = {{\left[ {{x_i},{y_i},{z_i}} \right]}^\T} \in {\mathbb{R}^3}}
\end{array}
\end{equation}
for $N$ as the number of points, and $\mathbf{c}_i$ as the 3D coordinates of the $i$th point, expressed in the camera frame.

\vspace{-0.2cm}
\subsection{Feature Sensorimotor Model}\label{section2b}
Let us denote the position of the robot's end-effector by $\mathbf{r} \in \mathbb{R}^{3}$.
As the dimension of the $3N$ observed points $\bar{\mathbf{c}}$ is typically very large, therefore, its direct use as a feedback signal for shape servocontrol is impractical.
Therefore, an efficient controller may typically require to use some form of dimension reduction technique.
To deal with this issue, we construct a feature vector $\mathbf s = \mathbf{f}_s(\bar{\mathbf{c}}): \mathbb R^{3N}\mapsto \mathbb{R}^{p}$, for $p \ll 3N$, to represent the geometric feedback $\bar{\mathbf{c}}$ of the object. 
This compact feedback-like signal will be used to design our automatic 3D shape controller.

For \emph{purely elastic} objects undergoing deformations, it is reasonable to model that the object configuration is only dependant on its potential energy $\mathcal P$  (thus, all inertial and viscous effects are neglected from our analysis \cite{navarro2017fourier}).
We model that $\mathcal P$ is fully determined by the feedback feature vector $\mathbf{s}$ and the robot's position $\mathbf{r}$ \cite{wakamatsu2004static}, i.e.: $\mathcal P = \mathcal P(\mathbf s,\mathbf{r})$.
In steady-state, the extremum expression satisfies $(\partial \mathcal P/\partial \mathbf{s})^\T = \mathbf a(\mathbf s,\mathbf r) = \mathbf{0}$ \cite{yu2021shape}.
We then define the following matrices:
\begin{equation}
\label{eq77}
{\frac{\partial^2 \mathcal P}{\partial \mathbf{s}^2} = \mathbf{G}(\mathbf{s},\mathbf{r})},\qquad
{\frac{\partial^2 \mathcal P}{{\partial{\mathbf{r}}} {\partial{\mathbf{s}}}} = \mathbf{K}(\mathbf{s},\mathbf{r})}
\end{equation}
which are useful to linearize the extremum equation (by using first-order Taylor's series expansion) as follows:
\begin{equation}
    \label{eq73}
    \mathbf{a}(\mathbf{s}+\Delta \mathbf{s},\mathbf{r}+\Delta \mathbf{r}) \approx \mathbf{a}(\mathbf{s},\mathbf{r})+\mathbf{G}(\mathbf{s},\mathbf{r})\Delta \mathbf{s} + \mathbf{K}(\mathbf{s},\mathbf{r}) \Delta \mathbf{r}
\end{equation}
for $\Delta \mathbf{s}$ and $\Delta \mathbf{r}$ as small changes.
Note that as $\mathbf{a}(\mathbf{s}+\Delta \mathbf{s},\mathbf{r}+\Delta \mathbf{r}) = \mathbf{a}(\mathbf{s},\mathbf{r}) = \mathbf{0}$ is satisfied, we can obtain the following motion model:
\begin{equation}
    \label{eq76}
    \Delta \mathbf{s} \approx -\mathbf{G}(\mathbf{s},\mathbf{r})^{-1} \mathbf{K}(\mathbf{s},\mathbf{r}) \Delta \mathbf{r} = \mathbf{J}(\mathbf{s},\mathbf{r}) \mathbf{u}
\end{equation}
where $\mathbf{J}(\mathbf{s},\mathbf{r})=-\mathbf{G}(\mathbf{s},\mathbf{r})^{-1} \mathbf{K}(\mathbf{s},\mathbf{r})\in \mathbb{R}^{p \times 3}$ represents the deformation Jacobian matrix (which depends on both the feature vector and robot position), and $\mathbf u$ represents the robot's motion control input.
This model can be expressed in an intuitive discrete-time form:
\begin{equation}
\label{eq55}
\Delta\mathbf{s}_{k+1} = \mathbf{J}_k(\mathbf{s}_k,\mathbf{r}_k) \mathbf{u}_k
\end{equation}
The deformation Jacobian matrix (DJM) $\mathbf J_k$ indicates how the robot's action $\mathbf{u}_k = \mathbf{r}_k - \mathbf{r}_{k-1}$ produces changes in the feedback features $\Delta\mathbf{s}_{k+1} = \mathbf{s}_{k+1}-\mathbf{s}_k$.
Clearly, the analytical computation of $\mathbf J_k$ requires knowledge of the physical properties and model of the elastic object and the vision system, which are difficult to obtain in practice.
Thus, numerical methods are often used to approximate this matrix in real-time, which enables to perform vision-guided manipulation tasks.

In this paper, we consider a robot manipulator whose control inputs represent velocity commands (here, modelled as the differential changes $\Delta\mathbf r$). 
It is assumed that $\Delta\mathbf r$ can be instantaneously executed without delay \cite{qi2021contour}.

\textbf{Problem statement.}
Design a vision-based control method to automatically manipulate a compliant object into a target 3D configuration, while simultaneously compensating for visual occlusions of the camera and estimating the deformation Jacobian matrix of the object-robot system.



\section{Shape Representation}\label{section3}
This section presents how online curve/surface fitting (least-squares minimization (LSM) \cite{abdi2007method}
and moving least squares (MLS) \cite{lancaster1981surfaces}) is combined with a parametric shape descriptor to compute a compact vector of feedback shape features.





\vspace{-0.3cm}
\subsection{LSM-Based Features}
\subsubsection{Centerline and Contour Extraction}\label{section3a}
Both are expressed as a parametric curve dependent on the normalized arc-length $0\le{\rho}\le{1}$. 
Then, the point can be represented as $\mathbf{c}_i=\mathbf{f}(\rho_i)$, with $\rho_i$ as the arc-length between the start point $\mathbf{c}_1$ and  $\mathbf{c}_i$, where $\rho_1 = 0$ and $\rho_N = 1$.
The fitting functional $\mathbf f(\cdot)$ is constructed as follows:
\begin{equation}
\label{eq2}
{\mathbf{f} \left( {{\rho}} \right) = \sum\limits_{j = 0}^n {{\mathbf{p} _j}{B_{j,n}}\left( {{\rho}} \right)}}
\end{equation}
where the vector $\mathbf{p}_j \in \mathbb{R}^3$ denotes the shape weights, $n\in \mathbb{N}^*$ specifies the fitting order, and the scalar $B_{j,n}(\rho)$ represents a parametric regression function, which may take various forms, such as: \cite{qi2020adaptive}
\begin{itemize}
	\item Polynominal parameterization \cite{smith1918standard}:
	\begin{equation}
	\label{eq3}
	{B_{j,n}}\left( {{\rho}} \right) = \rho^j
	\end{equation}
	
	\item Bernstein parameterization \cite{lorentz2013bernstein}:
	\begin{equation}
	\label{eq4}
	B_{j,n}\left( {{\rho}} \right) = C_n^j{\left( {1 - {\rho}} \right)^{n - j}}\rho^j
	\end{equation}
	where $C_b^j$ represents the binomial coefficient.
	
	\item Cox-deBoor parameterization \cite{schoenberg1973cardinal}:
	\begin{equation}
	\label{eq5}
	{B_{j,n}}\left( \rho \right) = \frac{1}{{n!}}\sum\limits_{l = 0}^{n - j} {\left( {{{\left( { - 1} \right)}^l}C_{n + 1}^l{{\left( {\rho + n - j - l} \right)}^n}} \right)} 
	\end{equation}
	
	\item Trigonometric parameterization \cite{powell1981approximation}:
	\begin{align}
	\label{eq6}
	{B_{j,n}}\left( {{\rho}} \right) = \left\{ {\begin{array}{*{20}{c}}
		1,&{j = 0} \\
		{\cos (\frac{j+1}{2} \rho)}, &{j > 0,\ j\ \rm{is}\ \rm{odd} } \\
		{\sin (\frac{j}{2}   \rho)}, &{j > 0,\ j\ \rm{is}\ \rm{even}}
		\end{array}} \right.
	\end{align}
\end{itemize}
From \eqref{eq1} and \eqref{eq2}, we can compute the following fitting cost function:
\begin{equation}
\label{eq66}
Q = {\left( {\mathbf{B}\mathbf{s} - \bar{\mathbf{c}}} \right)^\T}
	 \left( {\mathbf{B}\mathbf{s} - \bar{\mathbf{c}}} \right)
\end{equation}
for a ``tall'' regression-like matrix $\mathbf{B}$ constructed as:
\begin{align}
\label{eq65}
\mathbf{B} &= [\mathbf{B}_1^\T, \ldots, \mathbf{B}_N^\T]^\T 
\in \mathbb{R}^{3N \times 3(n+1)} \notag \\
\mathbf{B}_i &= [B_{0,n}(\rho_i),\ldots,B_{n,n}(\rho_i)] \otimes \mathbf{E}_3 \in \mathbb{R}^{3 \times 3(n+1)}
\end{align}
and $\mathbf{s} = [\mathbf{p}_0^\T, \ldots, \mathbf{p}_n^\T]^\T 
\in \mathbb{R}^{3(n+1)}$ as a compact feedback feature vector that represents the shape.
We seek to minimize \eqref{eq66} to obtain a feature vector $\mathbf{s}$ that closely approximates
$\bar{\mathbf c} \approx \mathbf B\mathbf s$.
The solution to \eqref{eq66} is:
\begin{equation}
\label{eq68}
\mathbf{s} = {\left( {{\mathbf{B}^\T}\mathbf{B}} \right)^{ - 1}}{\mathbf{B}^\T}\bar{\mathbf{c}}
\end{equation}
where it is assumed that $N\gg n+1$\footnote{In the following sections, $Q$ is used to generically represent different, albeit related, functions.}.




\subsubsection{Surface Extraction}\label{section3b}
The equation of the surface is defined as follows:
\begin{equation}
\label{eq19}
z= f(x,y) = \sum\limits_{j = 0}^{n_x} {\sum\limits_{l = 0}^{n_y} {{B_{j,n_x}}\left( {{x}} \right){B_{l,n_y}}\left( y \right){q_{jl}}} }
\end{equation}
where $n_x, n_y \in \mathbb{N}^*$ are the fitting order along with $x$ and $y$ direction, and $q_{jl} \in \mathbb{R}$ is the shape weight.
Same as with \eqref{eq66}, as fitting cost function is introduced:
\begin{equation}
\label{eq69}
Q = 
{\left( {\mathbf{D}\mathbf{s} - {\mathbf{z}}} \right)^\T}
 \left( {\mathbf{D}\mathbf{s} - {\mathbf{z}}} \right)
\end{equation}
for a ``augmented'' regression matrix $\mathbf{D}$ satisfying:
\begin{align}
    \label{eq80}
    \mathbf{D} &= [\mathbf{D}_1^\T,\ldots,\mathbf{D}_N^\T]^\T \in \mathbb{R}^{N \times (n_x+1)(n_y+1)} \notag \\ 
    \mathbf{D}_i &= [B_{0,n_x}(x_i)B_{0,n_y}(y_i), \ldots, B_{n_x,n_x}(x_i)B_{n_y,n_y}(y_i)] \notag \\
    \mathbf{D}_i^\T &\in \mathbb {R}^{(n_x+1)(n_y+1)}, \ for \ \ i=1,\ldots,N 
\end{align}
The depth vector is defiend as ${\mathbf{z}} = [z_1, \ldots, z_N]^\T \in \mathbb{R}^{N}$, and the feature vector is $\mathbf{s} = {\left[ {{q_{00}}, \ldots ,{q_{{n_x}{n_y}}}} \right]^\T} \in {\mathbb{R}^{\left( {{n_x} + 1} \right)\left( {{n_y} + 1} \right)}} $.
The solution to the minimization of \eqref{eq69} that approximates $\mathbf z \approx \mathbf D\mathbf s$ is as follows: 
\begin{equation}
\label{eq72}
\mathbf{s} = 
{\left( {{\mathbf{D}^\T}\mathbf{D}} \right)^{-1}}{\mathbf{D}^\T}{\mathbf{z}}
\end{equation}
For $N \gg (n_x+1)(n_y+1)$.



\begin{figure}[t]
	\centering
	\includegraphics[scale=0.21]{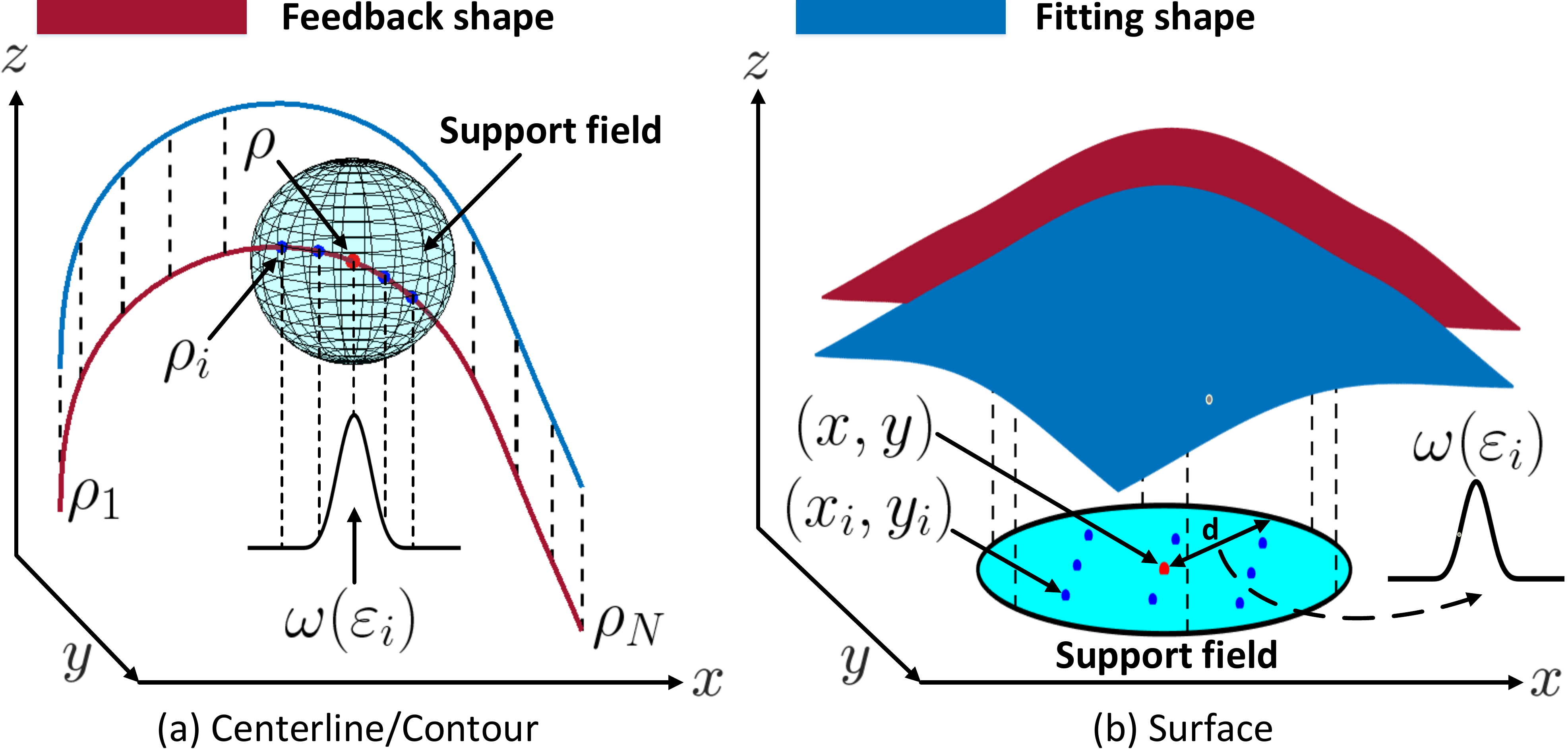}
	\vspace{-0.3cm}
	\caption{
		Scheme diagram of MLS.
		The weight $\omega (\varepsilon_i)$ is the square error between the fitted value and the given value.
		The fitting smoothness is regulated by adjusting the weight values.}
	\label{fig44}
\end{figure}


\subsection{MLS-based Feature Extraction}\label{section4}
Although LSM has an efficient one-step calculation, the weights of $B_{j,n}$ are the same all over the variable parameters (e.g., $\rho$ or $x,y$). 
Thus, to approximate complex curves/surfaces, the fitting order needs to be increased, which may lead to over-fitting problems.
To address this issue, MLS assumes that $\mathbf{s}$ is parameter-dependent, i.e., it changes with respect to $\rho$ (for centerline and contour), or to $(x,y)$ (for surface).
This enables MLS to represent complex shapes with a lower fitting order.
A support field \cite{tey2021moving} is introduced to ensure that the value of each weight is only influenced by data points within the support field.
A conceptual  diagram is given in Fig. \ref{fig44}.

\subsubsection{Centerline and Contour Extraction}\label{section4a}
The parametric equation with $\boldsymbol{\sigma}_j(\rho)$ is presented by referring to \eqref{eq2}:
\begin{equation}
\label{eq11}
{\mathbf{f} \left( {{\rho}} \right) 
	= \sum\limits_{j = 0}^n {{\boldsymbol{\sigma}_{j} (\rho)}{B_{j,n}}\left( {{\rho}} \right)}}
\end{equation}
where the parameter-dependent weight is denoted as $\boldsymbol{\sigma}_{j} (\rho) \in \mathbb{R}^3$.
The weighted square residual function is defined as:
\begin{equation}
\label{eq12}
Q(\rho)
= 
\sum\limits_{i = 1}^N \omega \left( \varepsilon_i  \right) \biggl\| {  \sum\limits_{j = 0}^n {{\boldsymbol{\sigma}_j}\left( \rho  \right){B_{j,n}}\left( {{\rho _i}} \right)} - {\mathbf{c}_i}} \biggr\|^2
\end{equation}
where $\varepsilon_i = |\rho - \rho_i| / d >  0$, and $d$ is the constant support field radius.
The scalar function $\omega(\varepsilon_i)$ is calculated as follows \cite{zhang2015measurement}:
\begin{equation}
\label{eq13}
\omega \left( \varepsilon_i \right) = \left\{ {\begin{array}{*{20}{c}}
	{\frac{2}{3} - 4\varepsilon_i^2 + 4\varepsilon_i^3},&{\varepsilon_i \le 0.5}\\
	{\frac{4}{3} - 4\varepsilon_i + 4\varepsilon_i^2 - \frac{4}{3}\varepsilon_i^3},&{0.5 < \varepsilon_i \le 1}\\
	0,&{\varepsilon_i > 1}
	\end{array}} \right.
\end{equation}
The function $\omega(\varepsilon_i)$ indicates the weight of $\rho$ relative to $\rho_i$, $\omega(\varepsilon_i)$ decreases as $\varepsilon_i$ increasing inside the support field.
When $\rho$ is outside the support field, $\omega(\varepsilon_i) = 0$.
MLS reduces into LSM when $\omega(\varepsilon_i)$ is constant.
The cost \eqref{eq12} can be equivalently expressed in matrix form as:
\begin{equation}
\label{eq14} 
Q(\rho) = 
\left( {\mathbf{B}\boldsymbol{\vartheta}(\rho) - \bar{\mathbf{c}}} \right)^\T
\mathbf{W}(\varepsilon)
\left( {\mathbf{B}\boldsymbol{\vartheta} (\rho) - \bar{\mathbf{c}}} \right)
\end{equation}
where $\boldsymbol{\vartheta} (\rho)$ and $\mathbf{W}(\varepsilon)$ are defined as follows:
\begin{align}
\label{eq15}
\boldsymbol{\vartheta}(\rho) &= [\boldsymbol{\sigma} _0^\T(\rho), \ldots, \boldsymbol{\sigma}_n^\T(\rho)]^\T 
\in \mathbb{R}^{3(n+1)}  \\
\mathbf{W} (\varepsilon)&= 
\diag ({ \omega( \varepsilon_1), \ldots,
		 \omega( \varepsilon_N)}) 
	     \otimes \mathbf{E}_3 
	     \in {\mathbb{R}^{3N \times 3N}}  \notag
\end{align}
The value of $\boldsymbol{\vartheta}(\rho)$ that minimizes \eqref{eq14} is computed as follows:
\begin{equation}
\label{eq17}
\boldsymbol{\vartheta}(\rho) = 
(\mathbf{B}^\T\mathbf{W}(\varepsilon)\mathbf{B})^{-1} \mathbf{B}^\T \mathbf{W}(\varepsilon)\bar{\mathbf{c}}
\end{equation}
By completing $N$ iterations along the parameter $\rho_1,...,\rho_N$, we can compute the augmented shape features as:
\begin{equation}
\label{eq18}
\boldsymbol{\Pi} = [\boldsymbol{\vartheta}(\rho_1), \ldots \boldsymbol{\vartheta}(\rho_N)] \in \mathbb{R}^{3(n+1) \times N}
\end{equation}
As the dimension of \eqref{eq18} is very large, it is impractical to use its components as a feedback signal for control. Thus, principal components analysis (PCA) \cite{zhu2021vision} is used to reduce the augmented structure \eqref{eq18} into a compact form $\widetilde{\boldsymbol{\Pi}} \in \mathbb{R}^{3(n+1) \times m}$ where $m \ll N$ by selecting the first $m$ most significant dimensions.
The feature vector $\mathbf{s} \in \mathbb{R}^{3m(n+1)}$ is computed by vectorizing the elements of $\widetilde{\boldsymbol{\Pi}}$.


\subsubsection{Surface Extraction}\label{section4b}
The equation of the surface is constructed as:
\begin{equation}
\label{eq61}
{f(x,y) = \sum\limits_{j = 0}^{n_x} {\sum\limits_{l = 0}^{n_y} {{B_{j,n_x}}\left( {{x}} \right){B_{l,n_y}}\left( y \right){\varpi_{jl}}(x,y)}}}
\end{equation}
where $\varpi_{jl}(x,y) \in \mathbb{R}$ is the parameter-dependent weight related to $(x,y)$.
Algorithm \ref{algorithm2} gives a pseudocode description of this method.
\begin{algorithm}[t] 
	\caption{Surface fitting procedure of MLS.} 
	\label{algorithm2} 
	\begin{algorithmic}[1] 
		\Require 
		$\bar{\mathbf{c}}$, $n_x$, $n_y$, $m$, and $d$;	
		\State Calculate node distance:
		\begin{equation*}
		\varepsilon_i = \sqrt{(x-x_i)^2 + (y-y_i)^2} / d
		\end{equation*}
	
		\State Construct the fitting cost function:
		\begin{equation*}
		    \label{eq82}
		    Q =
		    {\left( {\mathbf{D}\boldsymbol{\phi}\left( {x,y} \right) - \mathbf{z}} \right)^\T}\mathbf{W}\left( \varepsilon \right)
		    \left( {\mathbf{D}\boldsymbol{\phi}\left( {x,y} \right) - \mathbf{z}} \right)
		\end{equation*}
		for $\boldsymbol{\phi}(x,y)$ and $\mathbf{W}(\varepsilon)$ satisfying:
		\begin{align}
		    \label{eq83}
		    \boldsymbol{\phi} 
		    &= {[ {{\varpi_{00}}, \ldots ,{\varpi_{{n_x}{n_y}}}} ]^\T}  \notag \\
		    \mathbf{W}  &= \diag\left( {\omega \left( \varepsilon_1 \right), \ldots ,\omega \left( \varepsilon_N \right)} \right) 
		\end{align}
		
		\State Compute the structure $\boldsymbol{\phi}(x,y)$:
		\begin{equation*}
		\label{eq21}
		\boldsymbol{\phi} = ({{{\mathbf{D}^\T}\mathbf{W}\left( \varepsilon \right)\mathbf{D}}})^{-1}
		{{\mathbf{D}^\T}\mathbf{W}\left( \varepsilon \right)}{\mathbf{z}}
		\end{equation*}
		
		\State Use $x_i,y_i,i\in[1,N]$ to compute:
		\begin{equation}
		\boldsymbol{\Pi} = [\boldsymbol{\phi}(x_1,y_1), \ldots, \boldsymbol{\phi}(x_N,y_N)]
		\label{eq22}
		\end{equation}
		
		\State Use PCA to calculate 
		$\widetilde{\boldsymbol{\Pi}} $;
		
		\State Vectorize $\widetilde{\boldsymbol{\Pi}}$ to obtain 
		$\mathbf{s}$;
		\\ 
		\Return $\mathbf{s}$; 
	\end{algorithmic} 
\end{algorithm}



\begin{remark}
In addition to the proposed basis functions, there are other approaches that can be used to obtain similar results, e.g., Chebyshev polynomials \cite{mason2002chebyshev} and Legendre basis transformations \cite{farouki2000legendre}.
\end{remark}



\section{Shape Prediction Network}\label{section9}
\begin{figure}[t]
	\centering
	\includegraphics[scale=0.115]{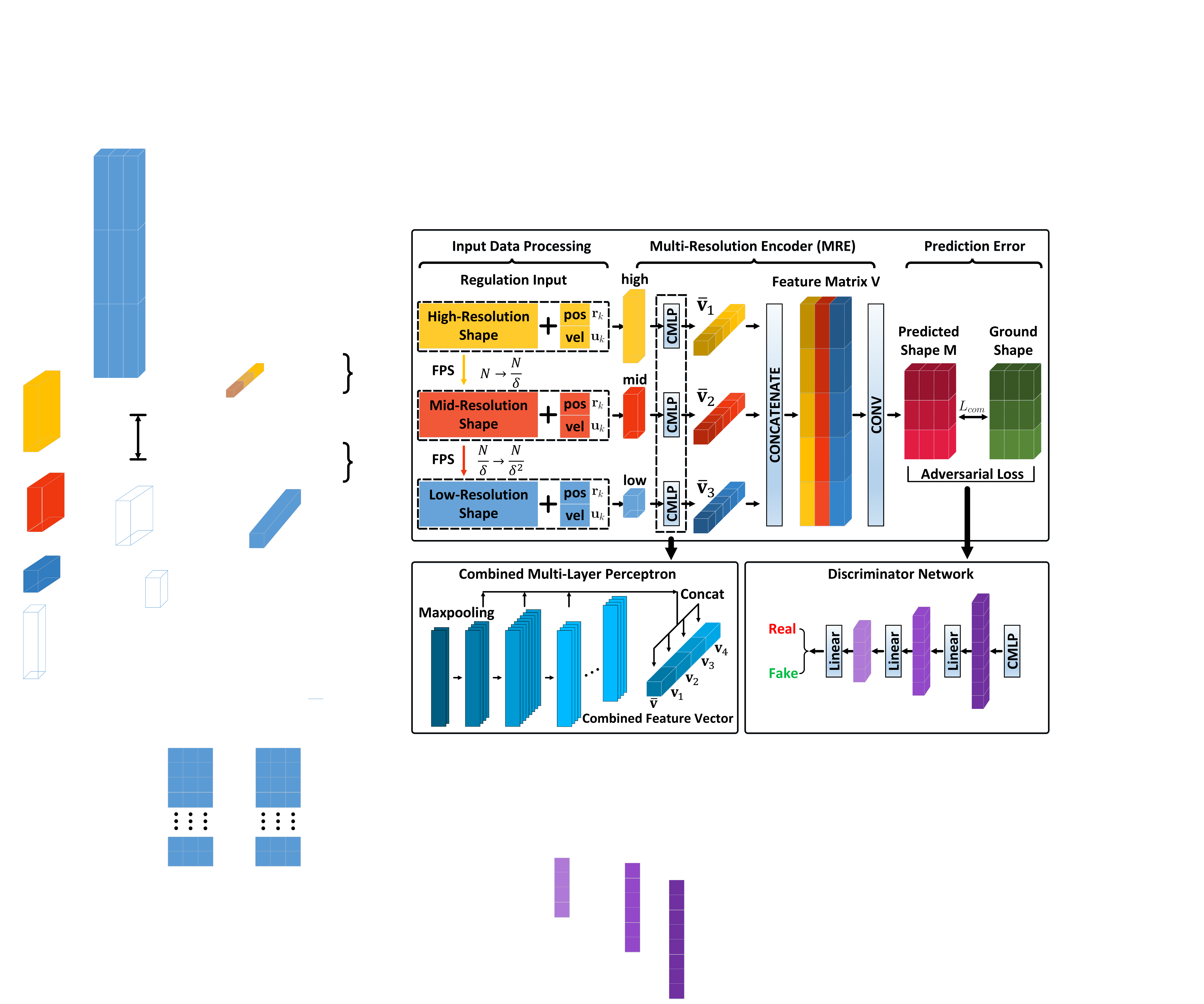}
	\vspace{-0.5cm}
	\caption{
	SPN predicts the next-moment shape $\bar{\mathbf{c}}_{k+1}$ by using the current-moment data in the case of occlusion, i.e., $\bar{\mathbf{c}}_{k} + \mathbf{r}_k + \mathbf{u}_k \xrightarrow{} \hat{\bar{\mathbf{c}}}_{k+1}$.
	FPS \cite{yan2020pointasnl} regulates $\bar{\mathbf{c}}_k$ (shown in yellow) uniformly to different scales (shown in red and blue).
	MRE stacks three-resolutions data together to form the total feature information, and obtains the predicted next-moment shape through convolution.
	DN is used to further improve the accuracy of the network.
	}
	\label{fig49}
	\vspace{-0.6cm}
\end{figure}

During the manipulation process, occlusions caused by obstacles or the robot itself may affect the integrity of observed shapes, and hence, the vector $\mathbf s$ cannot properly describe the object's configuration.
As a solution to this critical issue, in this paper we propose an occlusion compensation shape prediction network (SPN), which is composed of a regulation input (RI), a multi-resolution encoder (MRE) and a discriminator network (DN) \cite{huang2020pf}.
The proposed SPN utilizes the robot and object configurations and the active robot motions (i.e., $\bar{\mathbf{c}}_k$, $\mathbf{r}_k$, $\mathbf{u}_k$) as input to the network to predict the next instance  shape, here denoted by $\hat{\bar{\mathbf{c}}}_{k+1}$.
Fig. \ref{fig49} shows the overall architecture of the SPN.


\vspace{-0.3cm}
\subsection{Input Data Preprocessing}\label{section9c}
As the input data $\bar{\mathbf{c}}_k, \mathbf{r}_k, \mathbf{u}_k$ to the network have different sizes, they need to be rearranged into structures with unified dimensions.
To this end, $\bar{\mathbf{c}}_k$ is first rearranged into the matrix $\mathbf{T}^{high}_k=[\mathbf{c}_1,\mathbf{c}_2,\ldots,\mathbf{c}_N]^\T \in \mathbb{R}^{N \times 3}$.
Then, farthest point sampling (FPS) \cite{yan2020pointasnl} is used to downsample $\mathbf{T}^{high}_k$ to two resolutions $\mathbf{T}^{mid}_k\in\mathbb R^{\frac{N}{\delta} \times 3}$ and $\mathbf{T}^{low}_k\in\mathbb R^{\frac{N}{\delta^2} \times 3}$, for $\delta$ as the resolution scale.
Finally, the vectors $\mathbf{r}_k$ and $\mathbf{u}_k$ are rearranged into the matrices $\boldsymbol \Lambda _k^{high} = {\mathbf{I}_{N \times 1}} \otimes \mathbf{r}_k^\T\in\mathbb R^{N\times 3}$ and $\boldsymbol\Sigma _k^{high} = {\mathbf{I}_{N \times 1}} \otimes \mathbf{u}_k^\T\in\mathbb R^{N\times 3}$, which are similarly downsampled into mid and low resolutions as follows:
\begin{align}
    \boldsymbol\Lambda _k^{mid} &= {\mathbf{I}_{\frac{N}{\delta} \times 1}} \otimes \mathbf{r}_k^\T,\qquad
    \boldsymbol\Lambda _k^{low} = {\mathbf{I}_{\frac{N}{\delta^2} \times 1}} \otimes \mathbf{r}_k^\T , \\
    \boldsymbol\Sigma_k^{mid} &= {\mathbf{I}_{\frac{N}{\delta} \times 1}} \otimes \mathbf{u}_k^\T,\qquad
    \boldsymbol\Sigma _k^{low} = {\mathbf{I}_{\frac{N}{\delta^2} \times 1}} \otimes \mathbf{u}_k^\T
\end{align}
Thus, three different resolutions are generated for the network, high $\{\mathbf{T}_k^{high}, \boldsymbol\Lambda _k^{high}, \boldsymbol\Sigma_k^{high}\}$, mid $\{\mathbf{T}_k^{mid}, \boldsymbol\Lambda _k^{mid}, \boldsymbol\Sigma_k^{mid}\}$, and low $\{\mathbf{T}_k^{low}, \boldsymbol\Lambda _k^{low}, \boldsymbol\Sigma_k^{low}\}$, that provides a total input data of dimension $[N\times9, \frac{N}{\delta} \times 9, \frac{N}{\delta^2} \times 9]$.
$\mathbf{T}_k^{high}, \mathbf{T}_k^{mid}, \mathbf{T}_k^{low}$ are the geometric shapes of the object under different compression sizes specified by $\delta$.
Thus, these three terms can describe the potential feature structure of objects.
The proposed SPN aims to predict the object's shape that results from the robots actions, under the current object-robot configuration.
By using this multi-resolution data \{high, mid, low\}, the encoder can better learn the potential feature information of shapes.

\vspace{-0.3cm}
\subsection{Multi-Resolution Encoder}\label{section9a}
{
A combined multi-layer perceptron (CMLP) is the feature extractor of MRE, which uses the output of each layer in a MLP to form a multiple-dimensional feature vector. 
Traditional methods adopt the last layer of the MLP output as features, and do not consider the output of the intermediate layers, which leads to potentially losing important local information \cite{qi2017pointnet}.
CMLP enables to make good use of low-level and mid-level features that include useful intermediate-transition information \cite{huang2020pf}.
CMLP utilizes MLP to encode input data into multiple dimensions $[64,128,256,512,1024]$.
Then, we maxpool the output of the last four layers to construct a multiple-dimensional feature vector as follows:
\begin{align}
\mathbf{v}_{1} \in \mathbb{R}^{128}, \
\mathbf{v}_{2} \in \mathbb{R}^{256}, \ 
\mathbf{v}_{3} \in \mathbb{R}^{512}, \ 
\mathbf{v}_{4} \in \mathbb{R}^{1024}
\end{align}
The combined feature vector is constructed as: $\bar{\mathbf{v}}_i=[\mathbf{v}_{1}^\T,\mathbf{v}_{2}^\T,\mathbf{v}_{3}^\T,\mathbf{v}_{4}^\T]^\T \in \mathbb{R}^{1920}$.
Three independent CMLPs map three resolutions into three individual $\bar{\mathbf{v}}_i,$ for $i=1,2,3$.
Each $\bar{\mathbf{v}}_i$ represents the extracted potential information of each resolution.
Then, the augmented feature matrix $\mathbf{V}$ is generated by arranging its columns as $\mathbf{V}=[\bar{\mathbf{v}}_1, \bar{\mathbf{v}}_2, \bar{\mathbf{v}}_3] \in \mathbb{R}^{1920 \times 3}$, and further through 1D-convolution to obtain $\mathbf{M} \in \mathbb{R}^{N \times 3}$.
Finally, the predicted next-moment shape $\hat{\bar{\mathbf{c}}}_{k+1} \in \mathbb{R}^{3N}$ is obtained by vectorizing $\mathbf{M}$.
}
The prediction loss of MRE is:
\begin{equation}
\label{eq57}
{L_{mre}} = \left\| {\hat{\bar{\mathbf{c}}}_{k+1} - \bar{\mathbf{c}}_{k+1}} \right\|
\end{equation}
where $\bar{\mathbf{c}}_{k+1}$ represents the ground-truth next-moment shape in the training data-set.

\vspace{-0.3cm}
\subsection{Discriminator Network}\label{section9b}
Generative Adversarial Network (GAN) is chosen as DN to enhance the prediction accuracy.
For simplicity, we define $\Phi=\rm{MRE}()$ and $\Psi=\rm{DN}()$.
We define ($\bar{\mathbf{c}}_k, \mathbf{r}_k, \mathbf{u}_k$) as the $\mathcal{X}$ input into $\Phi$, while $\mathcal{Y}$ represents the true shape $\bar{\mathbf{c}}_{k+1}$.
$\Psi$ is a classification network with similar structure as CMLP, constituted by serial MLP layers $[128,256,512,1024]$ to distinguish the predicted shape $\Phi(\mathcal{X})$ and the real shape $\mathcal{Y}$.
We maxpool the last three layers of $\Psi$ to obtain feature vector $[256,512,1024]$.
Three feature vectors are concatenated into a latent vector $\mathbf{m} \in \mathbb{R}^{1792}$, and then passed through the fully-connected layers $[512, 256, 64, 1]$ followed by sigmoid-classifier to obtain the evaluation.
The adversarial loss is defined as follows:
\begin{equation}
\label{eq58}
{L_{adv}} = \sum\limits_{1 \le i \le v} {\log \left( {1 - \Psi\left( {{\beta_i}} \right)} \right)}  + \sum\limits_{1 \le j \le v} {\log \left( {\Psi\left( \Phi(\alpha_i) \right)} \right)}
\end{equation}
where $\alpha_i \in \mathcal{X}, \beta_i \in \mathcal{Y}, i=1,\ldots,v$, and $v$ is size of the dataset including $\mathcal{X}$ and $\mathcal{Y}$.
The total loss of SPN is:
\begin{equation}
\label{eq59}
L = \zeta_{mre} L_{mre} + \zeta_{adv} L_{adv}
\end{equation}
where $\zeta_{mre}$ and $\zeta_{adv}$ are the weights of $L_{mre}$ and $L_{adv}$, respectively, which satisfy the condition: $\zeta_{mre} + \zeta_{adv} = 1$.



\section{Receding-time Model Estimation}\label{section5}
In this paper, the objects are assumed to be manipulated by the robot slowly, thus $\mathbf{J}_k({\mathbf{s}_k,\mathbf{r}_k})$ is expected to change smoothly.
For ease of presentation, we omit the arguments of $\mathbf{J}_k({\mathbf{s}_k,\mathbf{r}_k})$ and denote it as as $\mathbf{J}_k$ from now on.
To estimate the DJM, three indicators are considered, viz., accuracy, smoothness, and singularity.
To this end, an optimization-based receding-time model (RTM) estimator is presented to estimate the changes of the Jacobian matrix, denoted by $\Delta \hat{\mathbf{\mathbf{J}}}_k = \hat{\mathbf{J}}_k - \hat{\mathbf{J}}_{k-1}$, which enables to monitor the estimation procedure.
$\Delta \hat{\mathbf{J}}_k$ can be obtained by considering the following three constraints:
\begin{itemize}
	\item 
	($Q_1$) Constraint of receding-time error \cite{mo2020automated}.
	As $\mathbf{J}_k$ depicts the relationship between $\Delta \mathbf{s}_{k+1}$ and $\mathbf{u}_k$ in a local range, thus we consider the accumulated error in $\eta$ past moments to ensure the estimation accuracy.
	$\eta$ is the receding window size.
	The receding-time error is given by:
	\begin{equation}
	\label{eq60}
	Q_1 = \sum\limits_{j = 1}^{\eta} {{{\gamma^{j}  \bigg\| {{\Delta\mathbf{s}_{k + 1 - j}} - {({\hat{\mathbf{J}}}_{k-1} + \Delta \hat{\mathbf{J}}_k)}{\mathbf{u}_{k - j}}} \bigg\|}^2}} 
	\end{equation}
	The sensitivity to noise can be improved by adjusting $\eta$, which helps to address the measurement fluctuations.
	$0<\gamma \le{1}$ is a constant forgetting factor giving less weight to the past observation data.

	
	\item 
	($Q_2$) Constraint of estimation smoothness \cite{mo2020automated}.
	As $\mathbf{J}_k$ is assumed to be smooth, thus $\Delta \hat{\mathbf{J}}_k$ should be estimated smoothly to avoid sudden large fluctuations, which can be achieved by minimizing the Frobenius norm of $\Delta \hat{\mathbf{J}}_k$:
	\begin{eqnarray}
	\label{eq74}
	Q_2 = \| \Delta \hat{\mathbf{J}}_k \|_F^2
	\end{eqnarray}

	\item 
	($Q_3$) Constraint of shape manipulability \cite{lynch2017modern}.
	It evaluates the feasibility of changing the object's shape under the current object-robot configuration:
	\begin{equation}
	\label{eq39}
	Q_3 = \bigg\| \frac{
		\lambda_{\max}((\hat{\mathbf{J}}_{k-1} + \Delta \hat{\mathbf{J}}_k)^\T(\hat{\mathbf{J}}_{k-1} + \Delta \hat{\mathbf{J}}_k))}
	              {\lambda_{\min}((\hat{\mathbf{J}}_{k-1} + \Delta \hat{\mathbf{J}}_k)^\T(\hat{\mathbf{J}}_{k-1} + \Delta \hat{\mathbf{J}}_k))}\bigg\|^2
	\end{equation}
	where $\lambda_{\max}$ and $\lambda_{\min}$ are the maximum and minimum eigenvalue, respectively, and $Q_3 \ge 1$.
	When $Q_3 = 1$, the object can deform isotropically in any direction. 
	A growing $Q_3$ indicates that the object is reaching  singular (non-manipulable) configuration.
	
\end{itemize}

Finally, the total weighted optimization index is given by:
\begin{align}
\label{eq48}
Q(\Delta \hat{\mathbf J}_k) = \mu_1 Q_1 + \mu_2 Q_2 + \mu_3 Q_3
\end{align}
where $\mu_i>0$ are the weights that specify the contribution of each constraint, and which satisfy $\mu_1 + \mu_2 + \mu_3 = 1$.
The index \eqref{eq48} is then solved by using numerical optimization tools (e.g., \emph{Matlab/fmincon} or \emph{Python/CasADi}) to obtain $\Delta \hat{\mathbf{J}}_k$ and thus, iteratively update the deformation Jacobian matrix as follows: $\hat{\mathbf{J}}_k = \hat{\mathbf{J}}_{k-1} + \Delta \hat{\mathbf{J}}_k$.

\section{Model Predictive Controller}\label{section6}
It is assumed that the matrix $\mathbf{J}_k$ has been accurately estimated at the time instant $k$ by the RTM, such that it satisfies $\Delta \mathbf{s}_{k+1} = {\hat{\mathbf{J}}_k}{\mathbf{u}_k}$.
Based on this model, we propose an MPC-based controller to derive the velocity inputs $\mathbf{u}_k$ for the robot, while taking saturation and workspace constraints into account.
Two vectors are defined as follow:
\begin{align} 
\label{eq31}
\bar{\mathbf{s}}_k 
&= {[{\mathbf{s}_{k + 1|k}^\T, \ldots, \mathbf{s}_{ k + h|k }^\T} ]^\T}
\in {\mathbb{R}^{ph}} \notag \\
\bar{\mathbf{u}}_k 
&= {[ {\mathbf{u}_{k|k}^\T, \ldots ,\mathbf{u}_{k + h - 1|k}^\T} ]^\T}
\in {\mathbb{R}^{3h}}
\end{align}
where $\bar{\mathbf{s}}_k$ and $\bar{\mathbf{u}}_k$ represent the predictions of $\mathbf{s}_k$ and $\mathbf{u}_k$ in the next $h$ periods, respectively.
The vectors $\mathbf{s}_{k+i|k}$ and $\mathbf{u}_{k+i|k}$ denote the $i$th predictions of $\mathbf{s}_k$ and $\mathbf{u}_k$ from the time instant $k$, where $\mathbf{s}_{k|k} = \mathbf{s}_k$, and $\mathbf{u}_{k|k} = \mathbf{u}_k$ must hold.
The prediction $\bar{\mathbf{s}}_k$ can be calculated from the estimated Jacobian matrix by noting that $\hat{\mathbf{\mathbf{J}}}_k\approx\hat{\mathbf{\mathbf{J}}}_{k+h}$ is satisfied during period $[k,k+h]$ (which is reasonable, given the regularity of the object). 
This way, the predictions are computed as follows: 
\begin{equation}
\label{eq33}
\begin{array}{*{20}{c}}
{\mathbf{s}_{k + j|k} = \mathbf{s}_k + \sum\limits_{i = 0}^{j - 1} {{{\hat{\mathbf{J}} }_k}\mathbf{u}_{k + i|k}}},&{j=1,\ldots,h}
\end{array}
\end{equation}
All predictions are then grouped and arranged into a single vector form:
\begin{align}
\label{eq35}
{\bar{\mathbf{s}}_k} &= \mathbf{A} \mathbf{s}_k + \boldsymbol{\Theta} {\bar{\mathbf{u}}_k}, \notag \\
\mathbf{A} &= \mathbf{I}_{h \times 1} \otimes \mathbf{E}_p \in {\mathbb{R} ^{ph \times p}}, \ 
\boldsymbol{\Theta} = \mathbf{L}_h \otimes \hat{\mathbf{J}}_k  \in {\mathbb{R}^{ph \times 3h}} 
\end{align}
In addition to $\bar{\mathbf{s}}_k$ and $\bar{\mathbf{u}}_k$, we define the constant sequence vector $\bar{\mathbf{s}}_k^*$ that represents the desired shape feature as:
\begin{equation}
\label{eq36}
\bar{\mathbf{s}}_k^* 
= {\left[ {\mathbf{s}_{k + 1|k}^{*\T}, \ldots ,\mathbf{s}_{k + h|k}^{*\T}} \right]^\T} \in {\mathbb{R}^{ph}}
\end{equation}
The cost function $Q\left( \bar{\mathbf{u}}_k \right)$ for the optimization of the control input is formulated as:
\begin{equation}
\label{eq37}
Q\left( \bar{\mathbf{u}}_k \right) 
= {\left( {{\bar{\mathbf{s}}_k} - {\bar{\mathbf{s}}_k^*}} \right)^\T} \boldsymbol{\Upsilon}_1 \left( {{\bar{\mathbf{s}}_k} - {\bar{\mathbf{s}}_k^*}} \right) + \bar{\mathbf{u}}_k^\T \boldsymbol{\Upsilon}_2 {\bar{\mathbf{u}}_k}
\end{equation}
where $\boldsymbol{\Upsilon}_1>0$ and $\boldsymbol{\Upsilon}_2>0$ are the weights for the error convergence rate and the smoothness of $\mathbf{u}_k$, respectively.
Two constraints are considered:
\begin{itemize}
	\item \emph{Saturation limits.}
	In practice, robots have limits on their achievable joint speeds. 
	These constraints are useful in soft object manipulation tasks to avoid damaging the object.
	Therefore, $\bar{\mathbf{u}}_k$ needs to be constrained:
	\begin{equation}
	\label{eq38}
	{\bar{\mathbf{u}}_{\min} \le \bar{\mathbf{u}}_k \le \bar{\mathbf{u}}_{\max}}
	\end{equation}
	where $\bar{\mathbf{u}}_{\min}$ and $\bar{\mathbf{u}}_{\max}$ are the constant lower and upper bounds, respectively.

	\item \emph{Workspace limits.}
	Robots are also often required to operate in a confined workspace to avoid colliding with the environment. 
	In soft object manipulation, this constraint is needed to avoid over-stretching or over-compressing the manipulated object. 
	To this end, the following constant bounds are introduced:
	\begin{equation}
	\label{eq40}
	\begin{array}{*{20}{c}}
	{\mathbf{r}_{k + i|k}^{\min } \le {\mathbf{r}_{k + i|k}} \le \mathbf{r}_{k + i|k}^{\max }},&{0 \le i \le h-1  }
	\end{array}
	\end{equation}
	Similarly as in \eqref{eq33}, the recursive structure of \eqref{eq40} can be obtained follows:
    \begin{equation}
	\label{eq41}
	{\boldsymbol{\Xi} _{\min }} \le \mathbf{C}{{\bar{\mathbf{u}} }_k} \le {\boldsymbol{\Xi} _{\max}}
	\end{equation}
	for $\boldsymbol{\Xi}_{\min},\boldsymbol{\Xi}_{\max} \in \mathbb{R}^{3h}$ and $\mathbf C\in\mathbb R^{3h\times 3h}$ defined as:
	\begin{align}
	\label{eq42}
	\boldsymbol{\Xi}_{\min} &= [
	({\mathbf{r}_{k|k}^{\min } - {\mathbf{r}_{k-1}}})^\T,...,
	({\mathbf{r}_{k + h - 1|k}^{\min } - {\mathbf{r}_{k-1}}})^\T]^\T 
	 \notag \\
	\boldsymbol{\Xi}_{\max} &= [
	({\mathbf{r}_{k|k}^{\max } - {\mathbf{r}_{k-1}}})^\T,...,
	({\mathbf{r}_{k + h - 1|k}^{\max } - {\mathbf{r}_{k-1}}})^\T]^\T 
	\notag \\
	\mathbf{C} &= \mathbf{L}_h \otimes \mathbf{E}_3 \in \mathbb{R}^{3h \times 3h}
	\end{align}
\end{itemize}
The quadratic optimization problem is formulated as follows:
\begin{equation}
\label{eq43}
\begin{array}{*{20}{c}}
\underset{\bar{\mathbf u}_k}{\min}~ 
Q(\bar{\mathbf{u}}_k) = \frac{1}{2}\bar{\mathbf{u}}_k^\T {\mathbf{H}} {\bar{\mathbf{u}}_k} + {\mathbf{q}^\T}{\bar{\mathbf{u}}_k}
\vspace{0.1cm}
\\
{\begin{array}{*{20}{c}}
	{\mathtt{s.t.}}&{\begin{array}{*{20}{c}}
		{{\bar{\mathbf{u}} }_{\min }} \le {{\bar{\mathbf{u}} }_k} \le {{\bar{\mathbf{u}} }_{\max }}
		\vspace{0.08cm}
		\\
		{\boldsymbol{\Xi} _{\min }} \le \mathbf{C}{{\bar{\mathbf{u}} }_k} \le {\boldsymbol{\Xi} _{\max }}
		\end{array}}
	\end{array}}
\end{array}
\end{equation}
where ${\mathbf{H}} = 2\left( {{\boldsymbol{\Theta} ^\T} \boldsymbol{\Upsilon}_1 \boldsymbol{\Theta}  + \boldsymbol{\Upsilon} _2} \right) \in {\mathbb{R}^{3h \times 3h}}$, $\mathbf{q}^\T = 2{\boldsymbol{\Omega} ^\T} \boldsymbol{\Upsilon}_1 \boldsymbol{\Theta}  \in {\mathbb{R} ^{1 \times 3h}}$ and  $\boldsymbol{\Omega}  = \mathbf{A} \mathbf{s}_k - \bar{\mathbf{s}}_k^* \in {\mathbb{R}^{ph}}$ are constant matrices.
Then, $\bar{\mathbf{u}}_k$ can be obtained by using a standard quadratic solver on \eqref{eq43}.
Finally, ${\mathbf{u}}_k$ is calculated by the receding horizon scheme:
\begin{equation}
\label{eq46}
\mathbf{u}_k = [\mathbf{E}_3,\mathbf{0},\ldots,\mathbf{0}]\bar{\mathbf{u}}_k
\end{equation}

\begin{figure*}[h]
	\centering
	\includegraphics[scale=0.27]{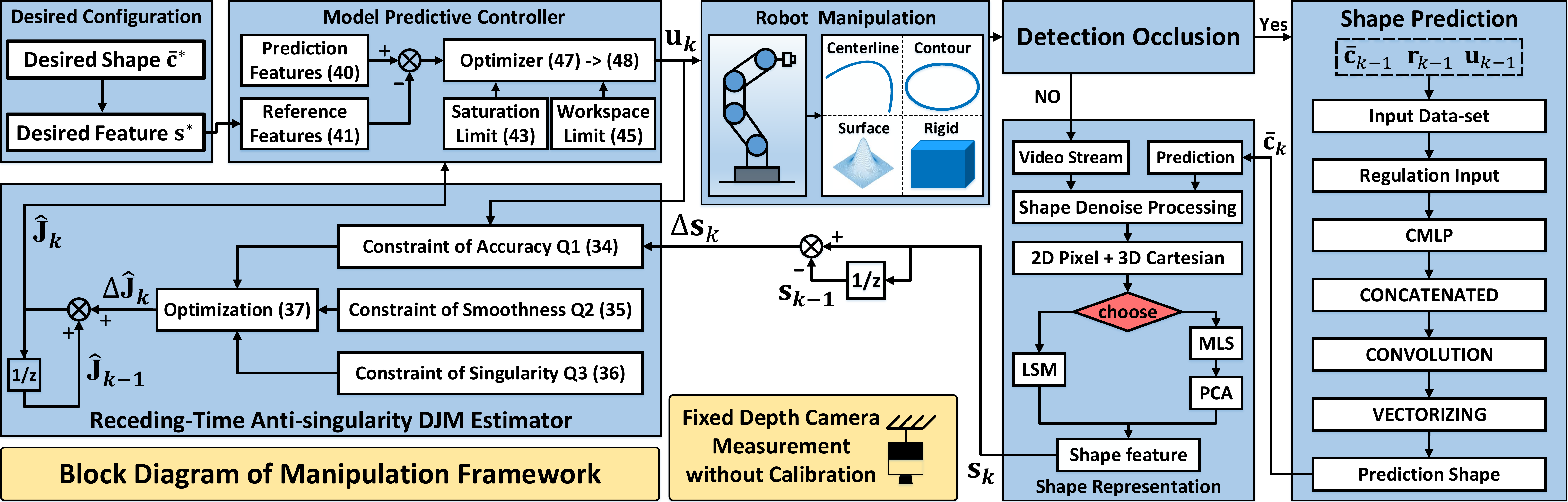}
	
	\vspace{-0.3cm}
	\caption{
	The block diagram of the proposed shape servoing framework, including representation, prediction, approximation, and manipulation within constraints.}
	\label{fig46}
	\vspace{-0.3cm}
\end{figure*}

Fig. \ref{fig46} presents the conceptual block diagram of the proposed framework.


\begin{remark}
The proposed MPC-based technique \eqref{eq43} computes the robot's shaping actions based on a performance objective and subject to system's constraints; This approach does not require the identification of the full analytical model of the deformable object. 
Its quadratic optimization form enables to integrate additional metrics into the problem, e.g., rising time and overshoot.
\end{remark}

		
		
		
		
		
		
		
		
		
		
		

\begin{figure}[h]
	\centering
	\subfloat[Centerline]
	{\includegraphics[width=2.1cm, height=1.65cm]{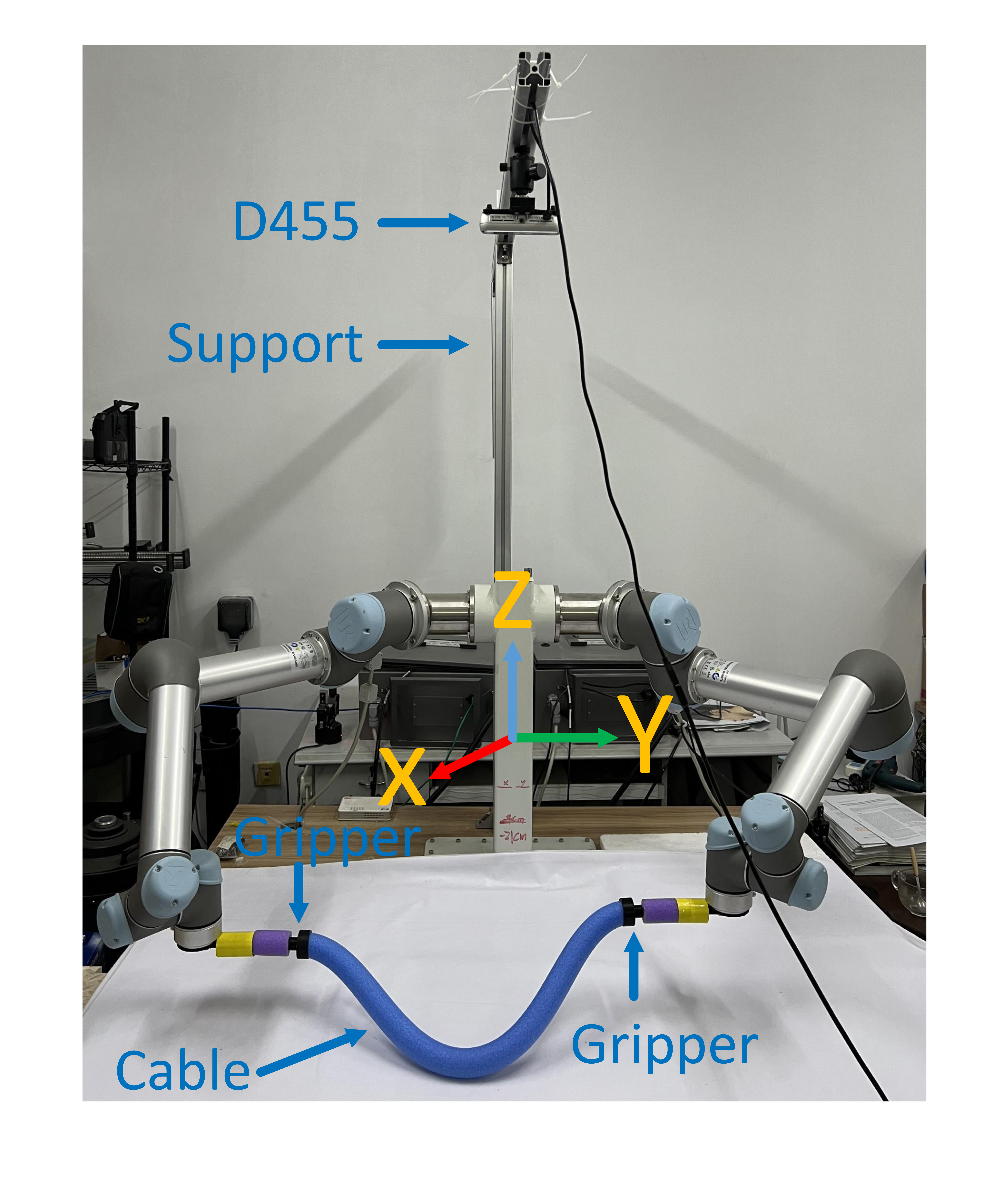}\label{fig41a1}}
	\subfloat[Contour]
	{\includegraphics[width=2.1cm, height=1.65cm]{fig41a2.pdf}\label{fig41a2}}
	\subfloat[Surface]
	{\includegraphics[width=2.1cm, height=1.65cm]{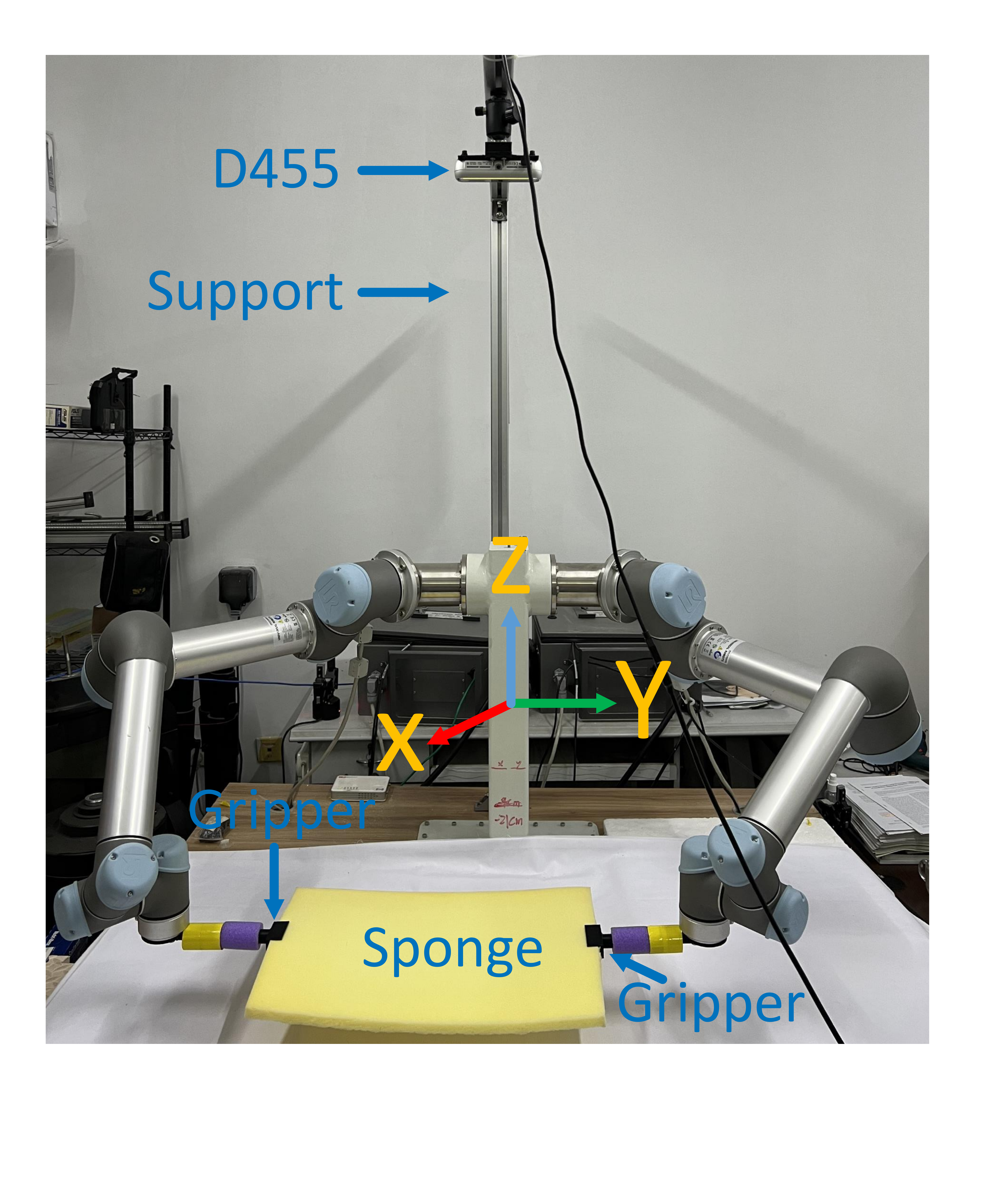}\label{fig41a3}}
	\subfloat[Plane]
	{\includegraphics[width=2.1cm, height=1.65cm]{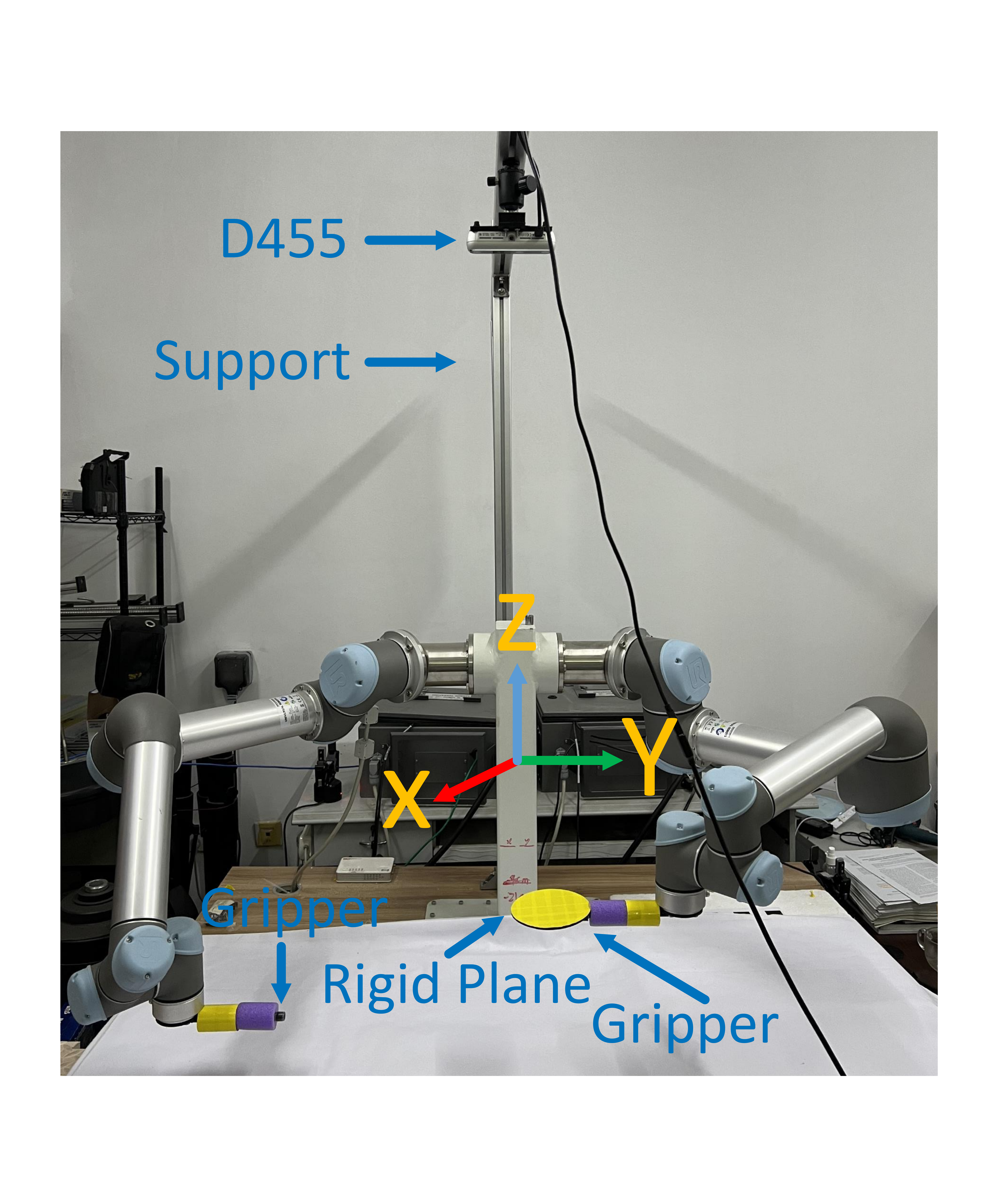}\label{fig41a4}}
	
	\vspace{-0.1cm}
	\caption{
	The experimental setup, including the objects (elastic and rigid), single-arm robot (UR5), and D455.}
	\label{fig27}
\end{figure}

\vspace{-0.5cm}
\section{Results}\label{section7}
\subsection{Experimental Setup}\label{section7a}
Vision-based manipulation experiments are conducted to validate our proposed framework.
The experimental platform used in our study includes a fixed D455 depth sensor, a UR5 robot manipulator, and various deformable objects shown in Fig. \ref{fig27}.
The depth sensor receives the video stream, from which it computes the 3D shapes by using the OpenCV and RealSense libraries.
In our experiments, only 3 DOF of the robot manipulator are considered, therefore, the control input
$\mathbf{u} = [u_x,u_y,u_z]^\T \in \mathbb{R}^3$ represents the linear velocity of the end-effector; A saturation limit of $|u_i| \le 0.01$ m/s, is applied for $i=x,y,z$.
The motion control algorithm is implemented on ROS/Python, which runs with a servo-control loop of
10 Hz.
A video of the conducted experiments can be downloaded from \url{https://github.com/JiamingQi-Tom/experiment_video/raw/master/paper4/video.mp4}

\begin{figure}[h]
	\centering
	\includegraphics[scale=0.34]{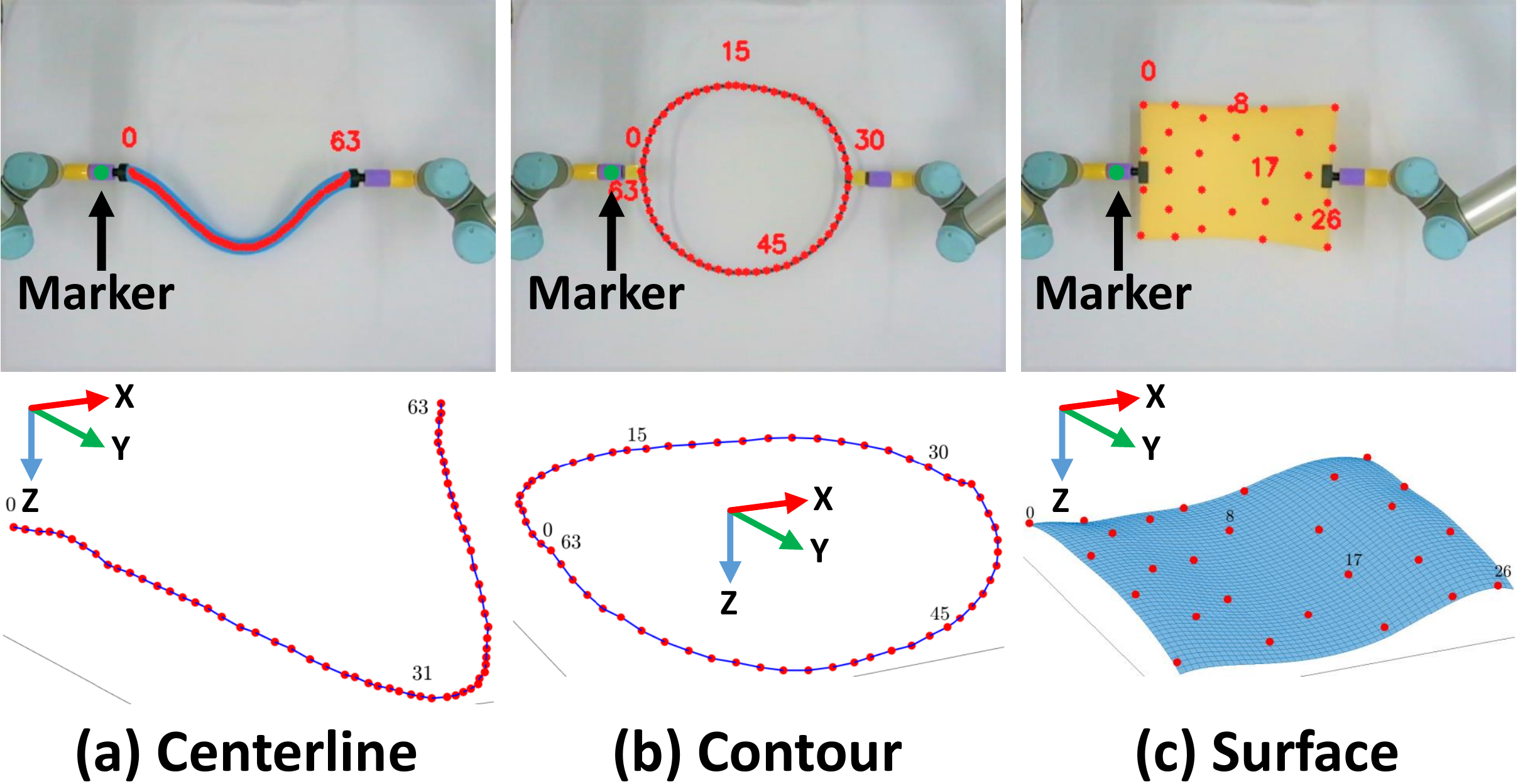}
	
	\vspace{-0.3cm}
	\caption{
	Shape extraction for centerline, contour and surface.
	The first row shows the 2-D image pixels, and the second row shows the 3D shapes.
	All shapes are fixed-sampled, equidistant, and ordered sorting.}
	\label{fig17}
\end{figure}

The proposed shape extraction algorithm is depicted in Fig. \ref{fig17}.
The RGB image from the camera is transformed into HSV and combined with mask processing to obtain a binary image.
$OpenCV/thinning$ is utilized with FPS to extract a fixed number of object points.
The point on the centerline closest to the gripper's green marker is chosen to sort the centerline along the cable.
Then we obtain the 3D shapes by using the RealSense sensor and checking its 2D pixels.
We adopt the method in \cite{qi2021contour} to compute the object's contour. 
A surface is obtained in the similar way to the centerline, i.e., by sorting points from top to bottom, and from left to right.
As 2D pixels and 3D points have a one-to-one correspondence in a depth camera, thus, our extraction method improves the robustness to measurement noise and is simpler than traditional point cloud processing algorithms.

%

\begin{figure}[h]
	\centering
	\includegraphics[scale=0.22]{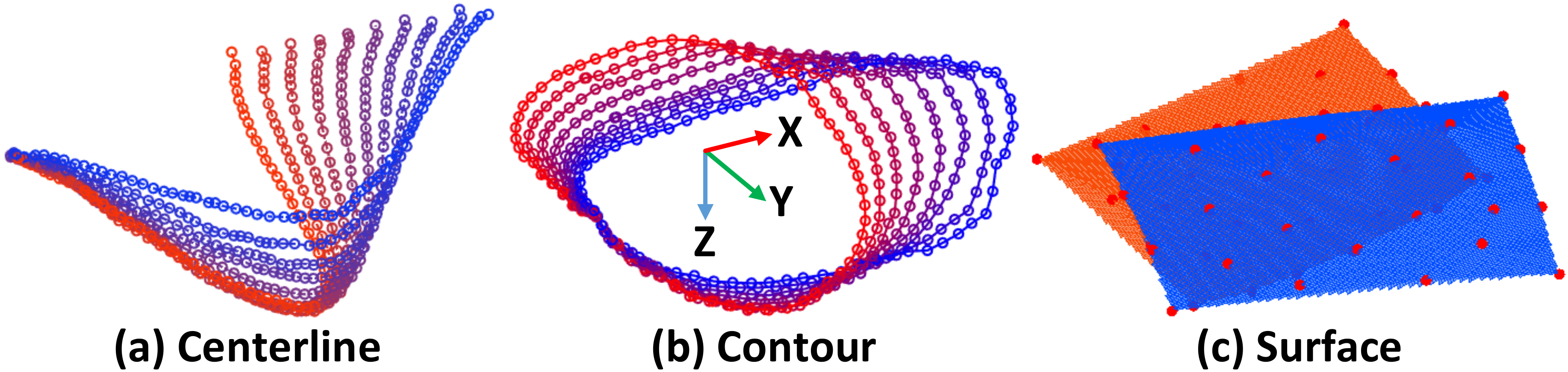}
	
	\vspace{-0.3cm}
	\caption{Shapes of various objects manipulated by UR5.}
	\label{fig62}
\end{figure}

\begin{figure}[h]
	\centering
	\subfloat[Centerline fitting]
	{\includegraphics[scale=0.23]{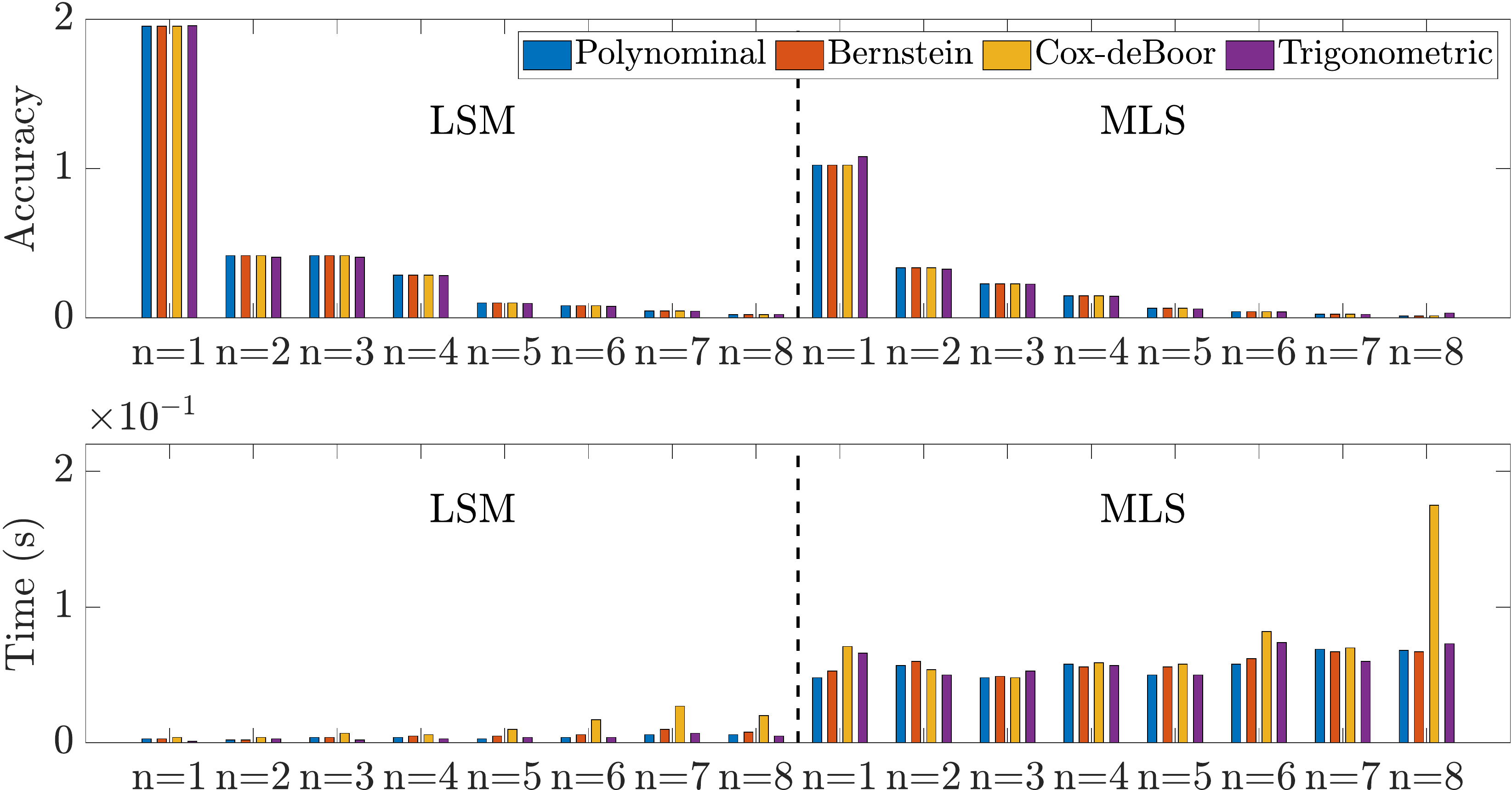}\label{fig16a}}
	
	\vspace{-0.35cm}
	\subfloat[Contour fitting]
	{\includegraphics[scale=0.23]{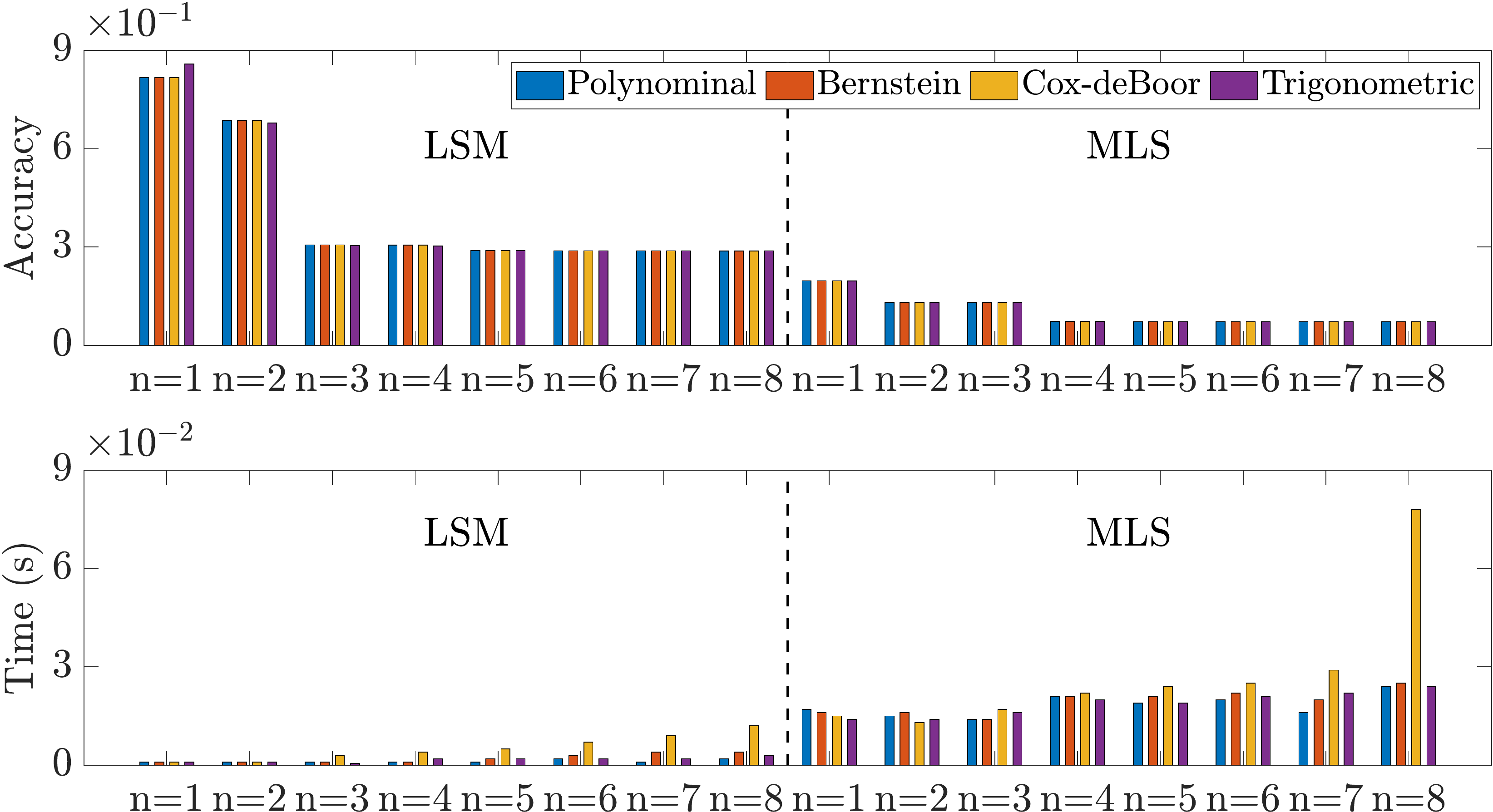}\label{fig16b}}
	
	\vspace{-0.35cm}
	\subfloat[Surface fitting]
	{\includegraphics[scale=0.23]{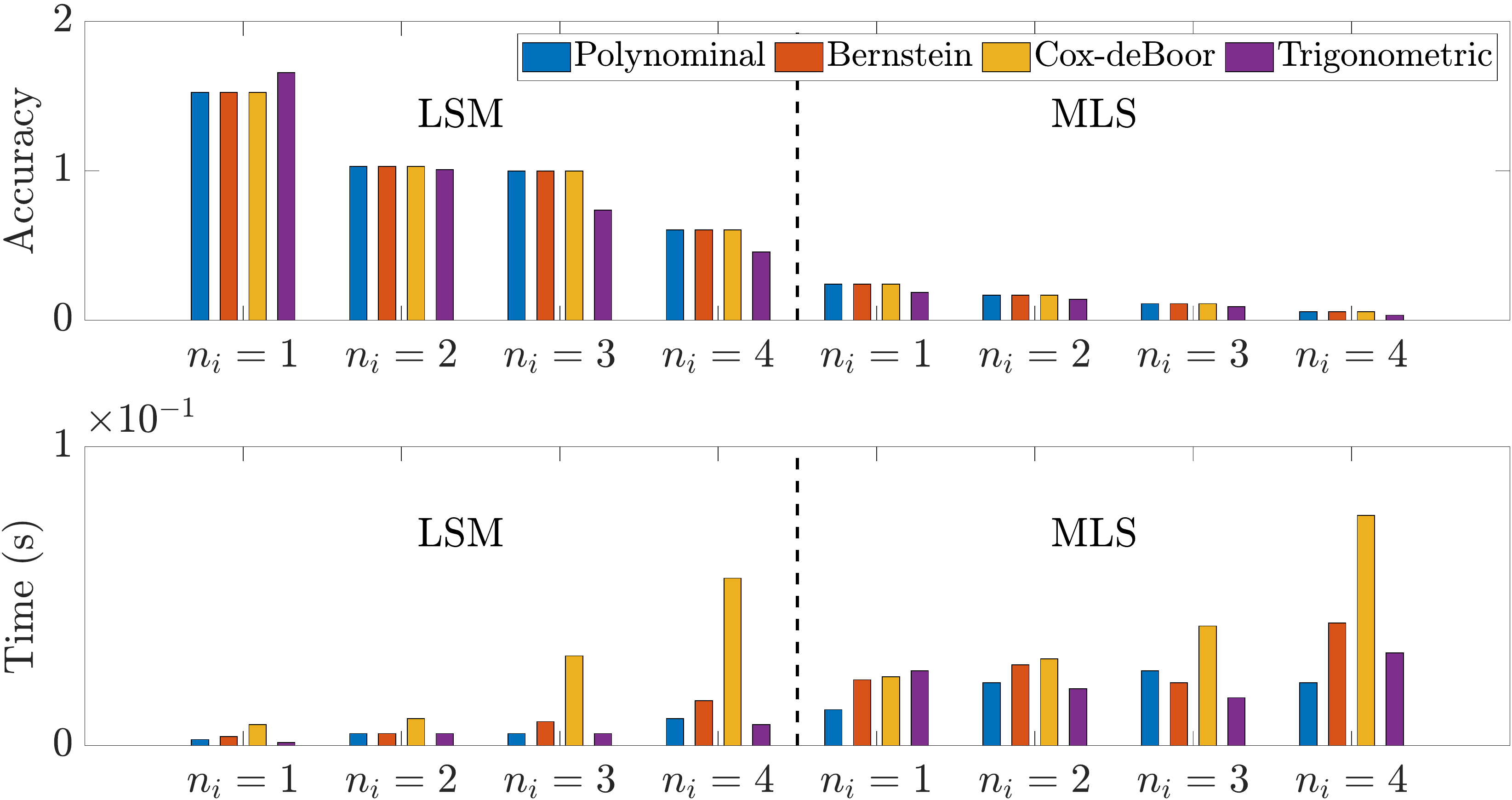}\label{fig16c}}
	
	\vspace{-0.2cm}
	\caption{
	Comparison of centerline, contour and surface fitting in accuracy and time-consuming among \eqref{eq3}\eqref{eq4}\eqref{eq5}\eqref{eq6} between LSM and MLS with $d=0.9$, $d=0.1$, and $d=0.2$, respectively.
	$n_i,i=x,y$ are the surface fitting order.
	}
	\label{fig16}
\end{figure}


\begin{table}[h]
	\caption{Fitting configurations. ``N/A'' stands for ``Not Applicable''.}
	\vspace{-0.3cm}
	\centering
	
	\begin{tabular}{c|ccc}
		\specialrule{0em}{1pt}{1pt}
		\toprule[2pt]
		
		\specialrule{0em}{1pt}{1pt}
		& Centerline  & Contour  & Surface\ / Rigid   \\ \hline 
		
		\specialrule{0em}{1pt}{1pt}
		Method  & LSM  & MLS & MLS  \\   
		
		\specialrule{0em}{1pt}{1pt}
		Support Radius  & N/A & $d=0.2$ & $d=0.2$  \\   
		
		\specialrule{0em}{1pt}{1pt}
		PCA  & N/A & $m=1$ & $m=1$  \\   
		
		\specialrule{0em}{1pt}{1pt}
		Fitting order   & $n=5$ & $n=4$ & $n_x=n_y=2$  \\     
		
		\specialrule{0em}{1pt}{1pt}
		Basis-Function  & Bernstein & Trigonometric & Polynominal  \\

		\specialrule{0em}{1pt}{1pt}
		Numbers  & $N=64$ & $N=64$ & $N=32$  \\ 
		\toprule[2pt]
	\end{tabular}
	\label{table3}
\end{table}

\vspace{-0.3cm}
\subsection{Online Fitting of the Parametric Shape Representation}
\label{section7c}
In this section, ten thousand samples of centerlines, contours and surfaces with $N=64, 64, 32$, respectively, are collected by commanding the robot to manipulate the objects, whose configuration is then captured by a depth sensor.
Such shaping actions are shown in Fig. \ref{fig62}, and visualized in the accompanying multimedia attachment. 
This data is used to evaluate the performance (viz. its accuracy and computation time) of our representation framework.
For that, we calculate the average error between the feedback shape $\bar{\mathbf{c}}$ and the reconstructed shape $\hat{\bar{\mathbf{c}}}$ as follows $\mean(\sum {\left\| {{{\bar{\mathbf{c}} }_i} - {{\hat{\bar{\mathbf{c}} } }_i}} \right\|})$.
The computation time is defined as the average of the overall processing time of all sample data among each method.

Fig. \ref{fig16} shows that the larger the scalars $n,n_x,n_y$ are, the better the fitting accuracy of LSM and MLS is.
MLS fits better than LSM under the same condition, as MLS calculates the independent weight while LSM assumes that each node has the same weight.
MLS works better in fitting contour because the parametric curve may not be continuous in the end corner, thus, the equal weight assumption of LSM is not suitable here.
As the number of data points for surface is $N=32$, it does not satisfy the condition $N \gg (n_x + 1)(n_y + 1)$ for higher order fitting models (e.g., $n_x=n_y \ge 5$), thus, we only use $n_x=n_y \le 4$.
The results show that MLS performs better than LSM in the surface representation; Interestingly, MLS obtains satisfactory performance even with $n_x=n_y=1$.
Fig. \ref{fig16} also shows that larger $n,n_x,n_y$ will also increase the computation time.
MLS has a more noticeable increase, as it calculates the weights of all nodes while LSM calculates them once.
The trigonometric approach is the fastest, polynomial and Bernstein follow, while Cox-deBoor is the slowest with the most iterative operations.
The above analysis verifies the effectiveness of the proposed extraction framework, which can represent objects with a low-dimensional feature.
Details of the curve fitting configuration is given in Table. \ref{table3}.



\begin{figure}[h]
	\centering
	\subfloat[Centerline]
	{\includegraphics[scale=0.45]{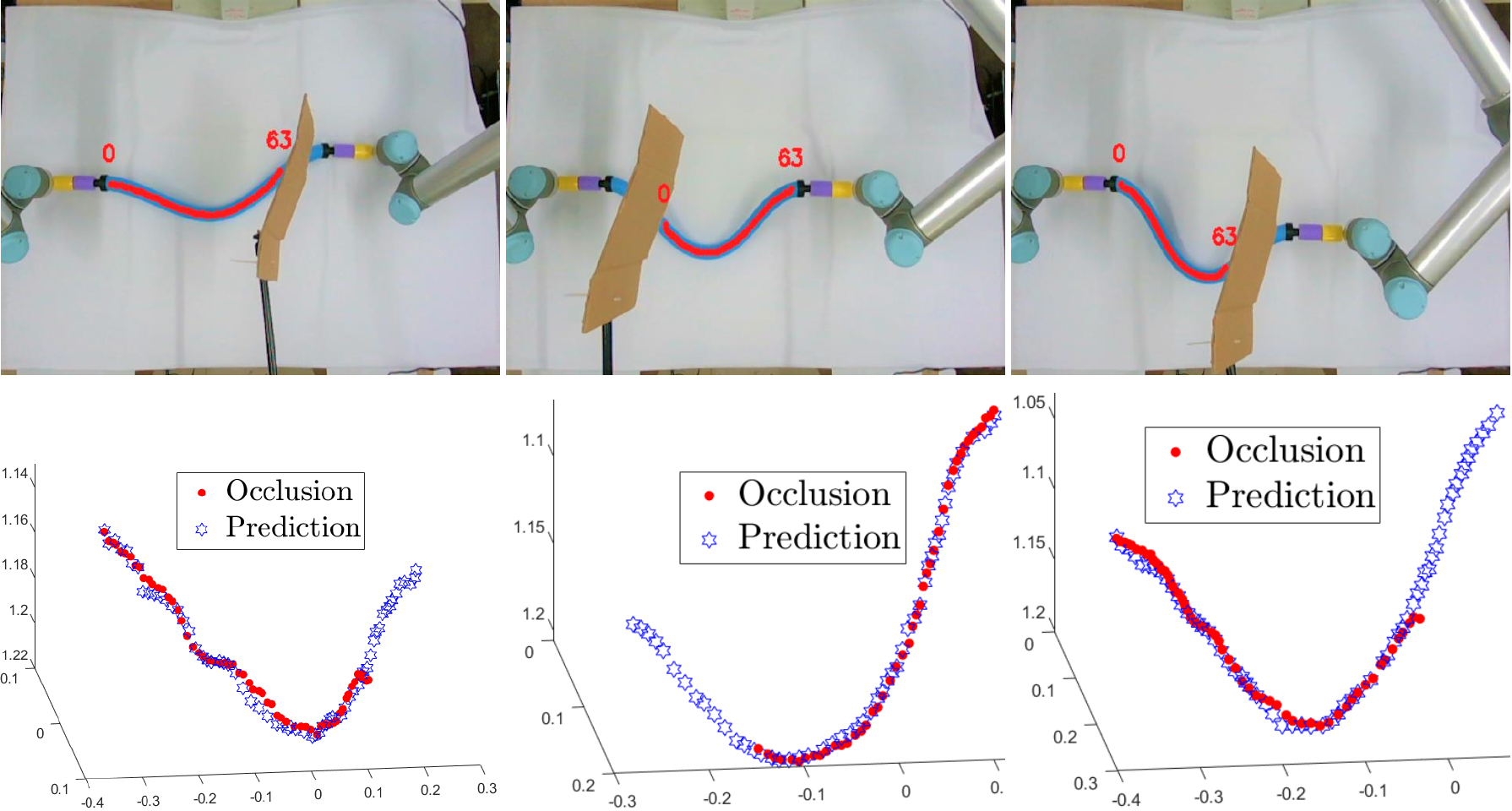}}
	
	\vspace{-0.2cm}
	\subfloat[Contour]
	{\includegraphics[scale=0.45]{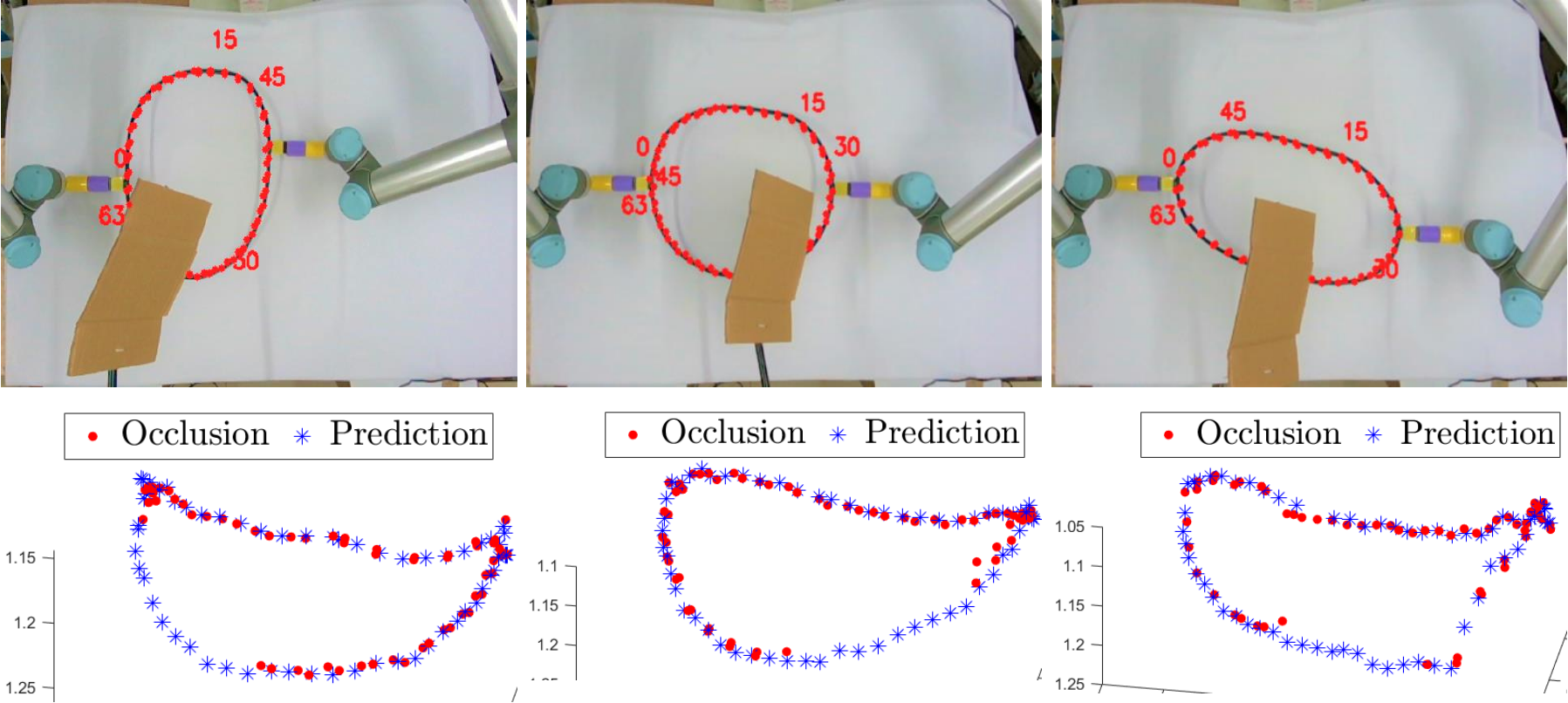}}
	
	\vspace{-0.2cm}
	\subfloat[Surface]
	{\includegraphics[scale=0.45]{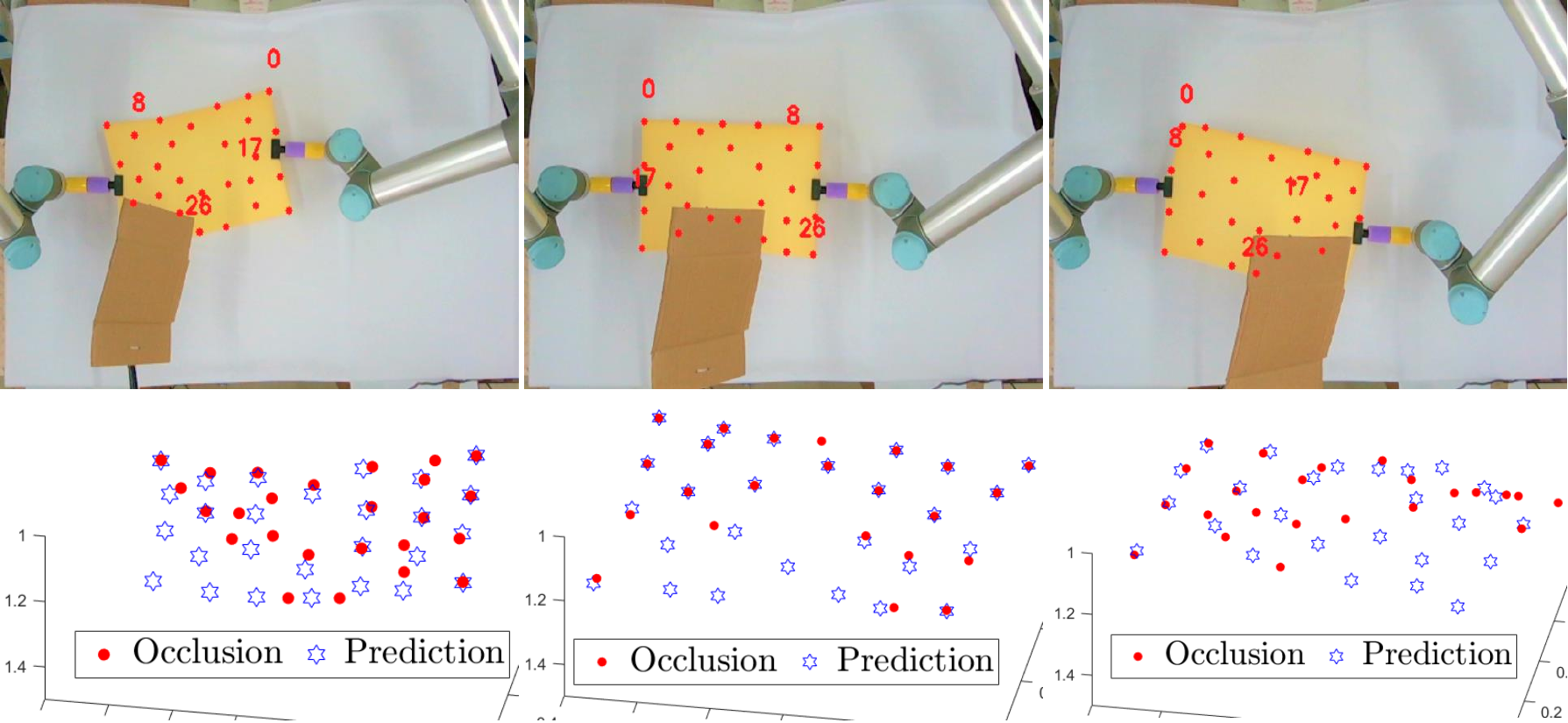}}
	
	\vspace{-0.2cm}
	\caption{
	Validation of SPN among three shape configurations with moving the obstacle manually.
	The RGB image describes the occluded case.
	Red dot in the second row gives the feedback shape within the occlusion at the current instant, and blue ones are the predicted shape according to the past data.
	}
	
	\label{fig18}
\end{figure}

\subsection{Occlusion-Robust Prediction of Object Shapes}
\label{section7e}
The data collected in Section \ref{section7c} is used to evaluate our proposed SPN.
80\% of the data is used as the training set, and the remaining 20\% of the data is used to test the trained network.
We set $\delta=2$ for the centerline, contour, and surface parametrization.
SPN is built using PyTorch and trained by an ADAM optimizer with a batch size of 500, and the initial learning rate set to 0.0001.
RELU activation and batch normalization are adopted to improve the network's performance.
In this validation test, the robot deforms the objects with small babbling motions while a cardboard sheet covers parts of objects.
Fig. \ref{fig18} shows that SPN can predict and provide relatively complete shapes for the three types of manipulated objects.
The accompanying video demonstrates the performance of this method.

\begin{figure}[h]
	\centering
	\subfloat[Centerline DJM Estimation]
	{\includegraphics[scale=0.24]{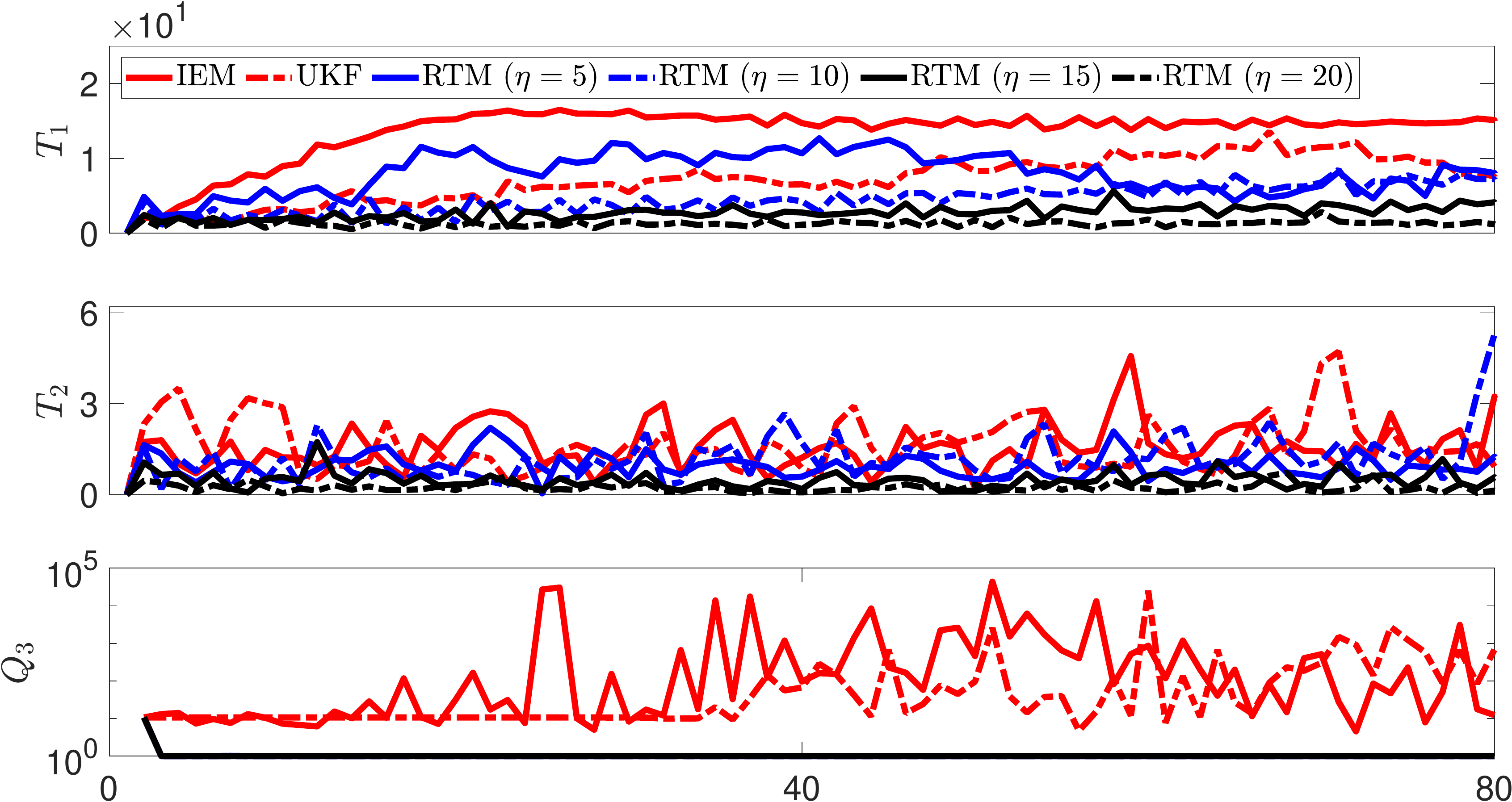}}
	\vspace{-0.4cm}
	
	\subfloat[Contour DJM Estimation]
	{\includegraphics[scale=0.24]{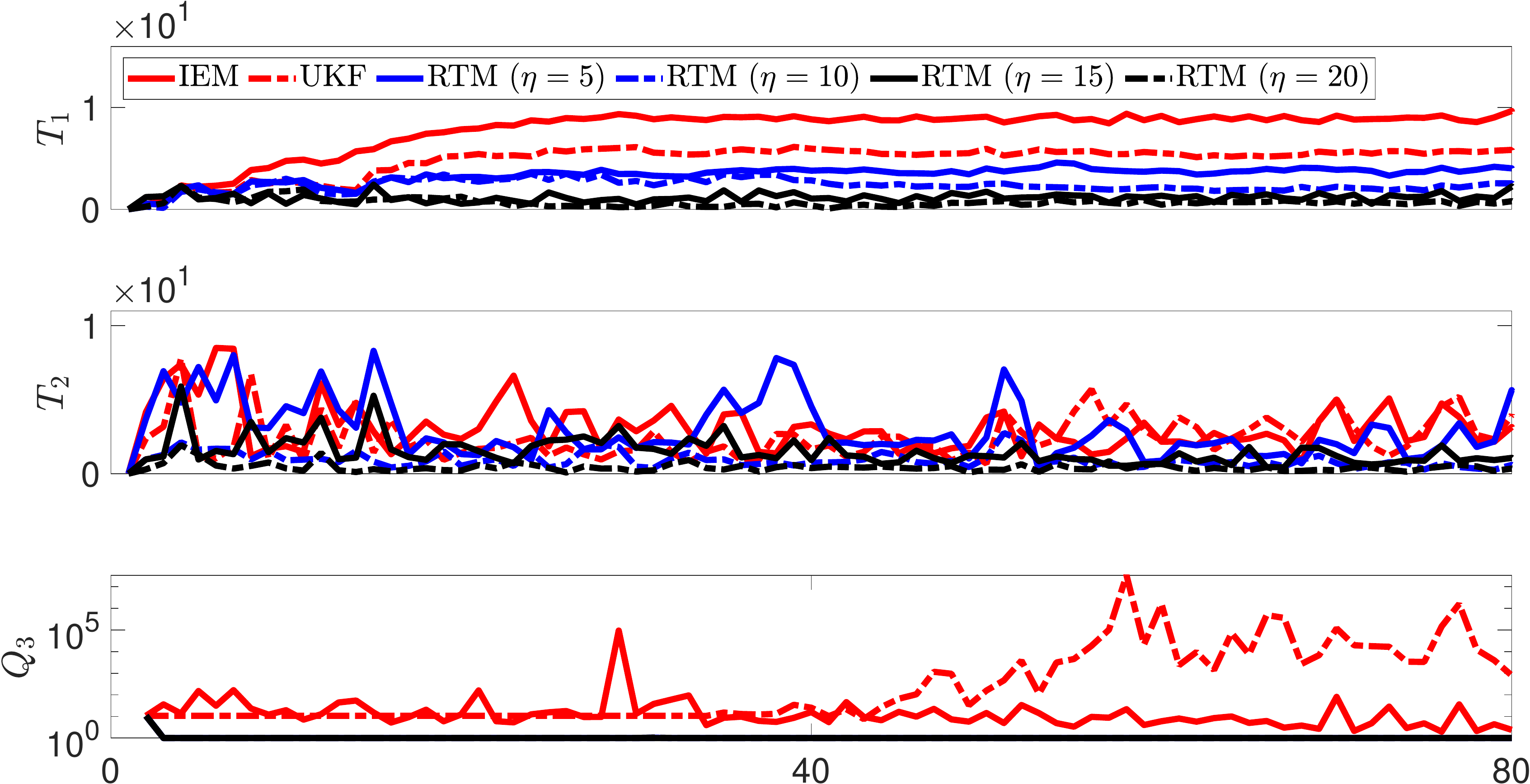}}
	\vspace{-0.4cm}
	
	\subfloat[Surface DJM Estimation]
	{\includegraphics[scale=0.24]{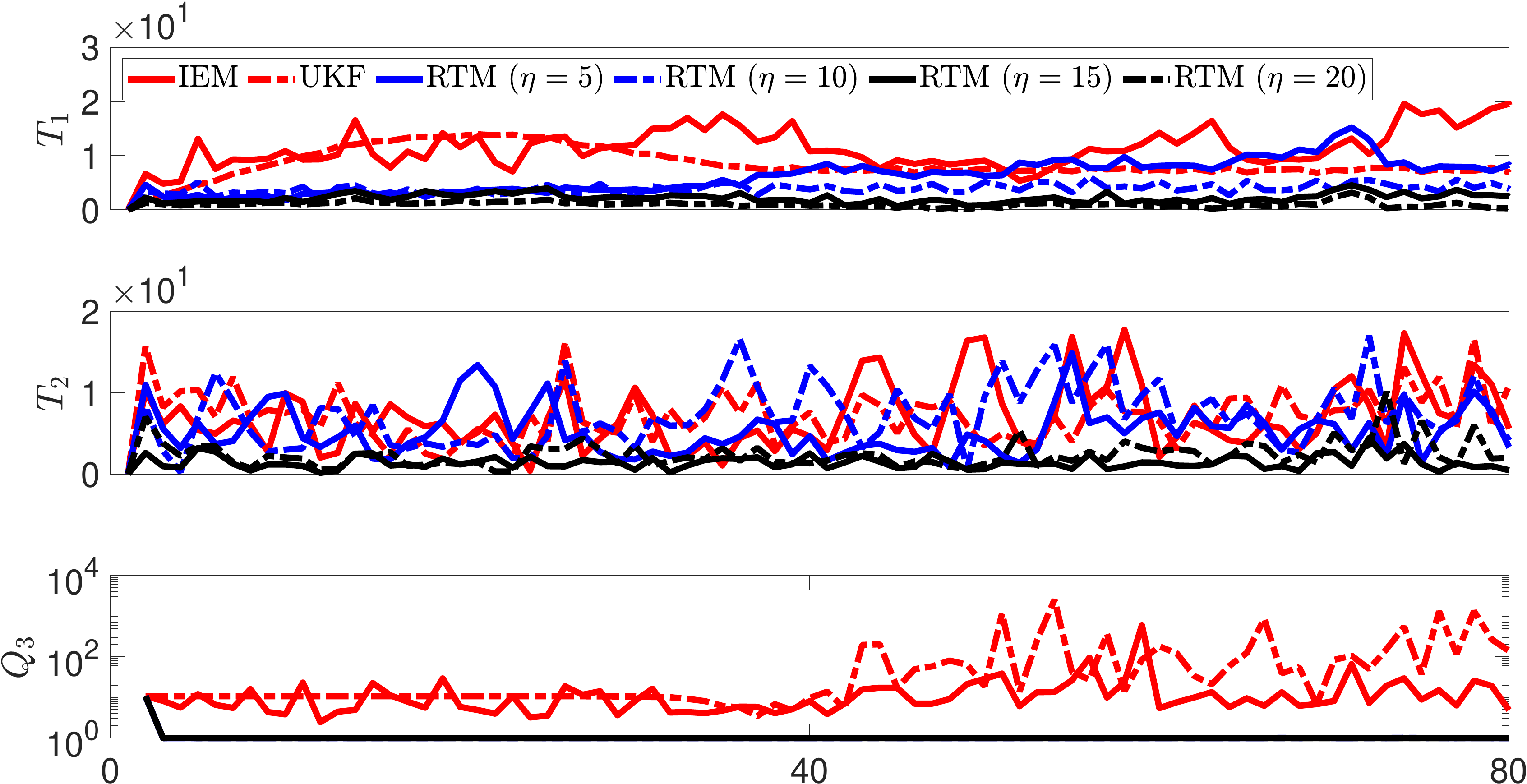}}
	
	\vspace{-0.1cm}
	\caption{
		Curves of $T_1, T_2$, and $Q_3$, reviewing the accuracy, smoothness, and singularity of DJM, respectively.
		$T_1$ represents the feature estimation error, and $T_2$ depicts the estimation differential error. 
		Besides, $\mu_1 = 0.8,\mu_2 = 0.1,\mu_3 = 0.1$.}
	\label{fig23}
	\vspace{-0.3cm}
\end{figure}

\subsection{Estimation of the Sensorimotor Model}\label{section7d}

This section aims to evaluate RTM \eqref{eq48} that approximates the deformation Jacobian matrix; The performance of the RTM estimator is compared with the interaction matrix estimator (IEM) in \cite{zhu2021vision}, and the unscented Kalman filter (UKF) in \cite{qi2020adaptive}.
To this end, we introduce two metrics, i.e., $T_1$ and $T_2$, to quantitatively compare performances of these algorithms:
\begin{align}
\label{eq24}
\begin{array}{*{20}{c}}
{{T_1} = \| {\widehat{{\mathbf{s}}}_{k+1} - {\mathbf{s}}_{k+1}} \|},&{{T_2} = \| {{\Delta \mathbf{s}_{k+1}} - {{\hat{\mathbf{J}}}_k}\mathbf{u}_k} \|}
\end{array}
\end{align}
where $\widehat{\mathbf{s}}_{k+1} = \widehat{\mathbf{s}}_{k} + \hat{\mathbf{J}}_k \mathbf{u}_k$ is the approximated shape feature that is computed based on the control actions.
Fig. \ref{fig23} shows that RTM with $\eta=20$ provides the best performance for $T_1$ among the methods; This means that constraint $Q_1$ \eqref{eq60} enables RTM to learn from past data by adjusting $\eta$.
Small values of $T_2$ reflect that RTM accurately predicts the differential changes induced by the DJM; This means that constraint $Q_2$ \eqref{eq74} helps to compute a smooth matrix $\hat{\mathbf{J}}_k$.
As the proposed method incorporates the constraint $Q_3$ \eqref{eq39}, thus, RTM can prevent singularities in the estimation of the Jacobian matrix, while IEM and UKF are prone to reach ill-conditioned estimations.
We use RTM with $\eta=10$ in the following sections.

\begin{figure*}[!htb]
\vspace{-0.3cm}
	\centering
	\subfloat[Exp1 / Centerline] 
	{\includegraphics[scale=0.34]{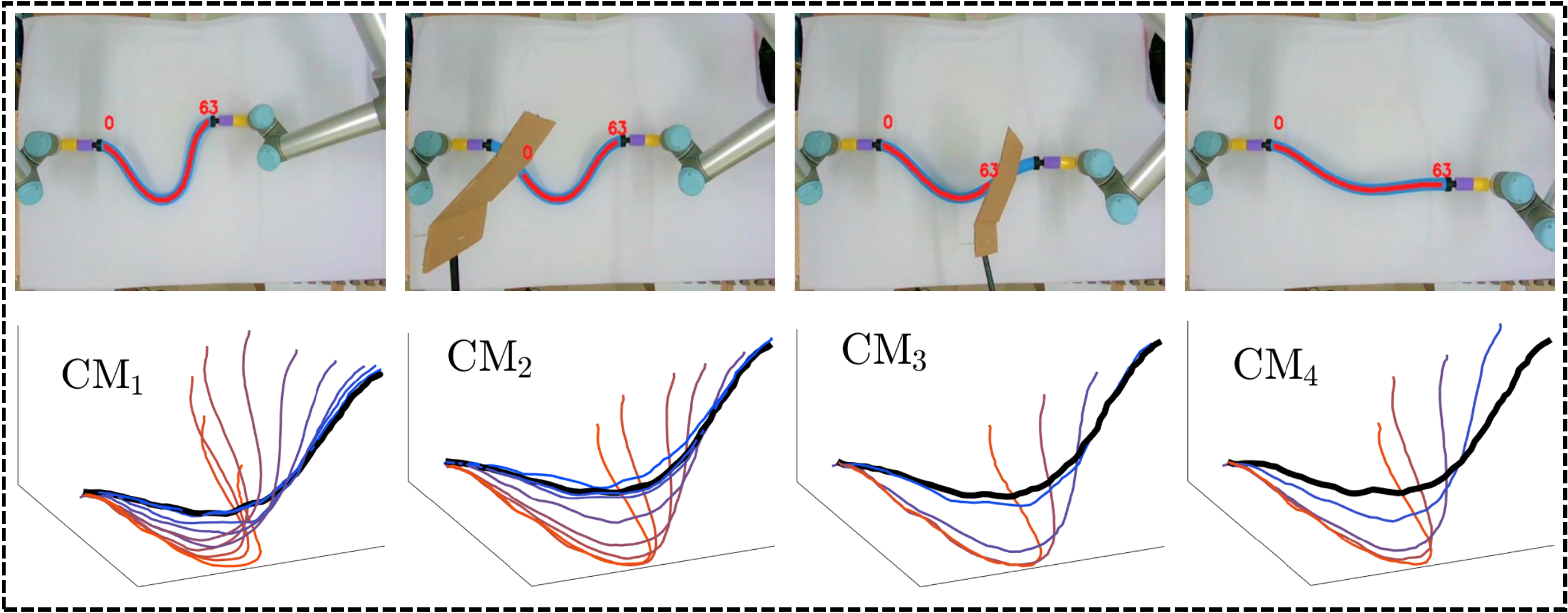}\label{fig63a}}
	\hspace{0.07cm}
	\subfloat[Exp2 / Contour]
	{\includegraphics[scale=0.34]{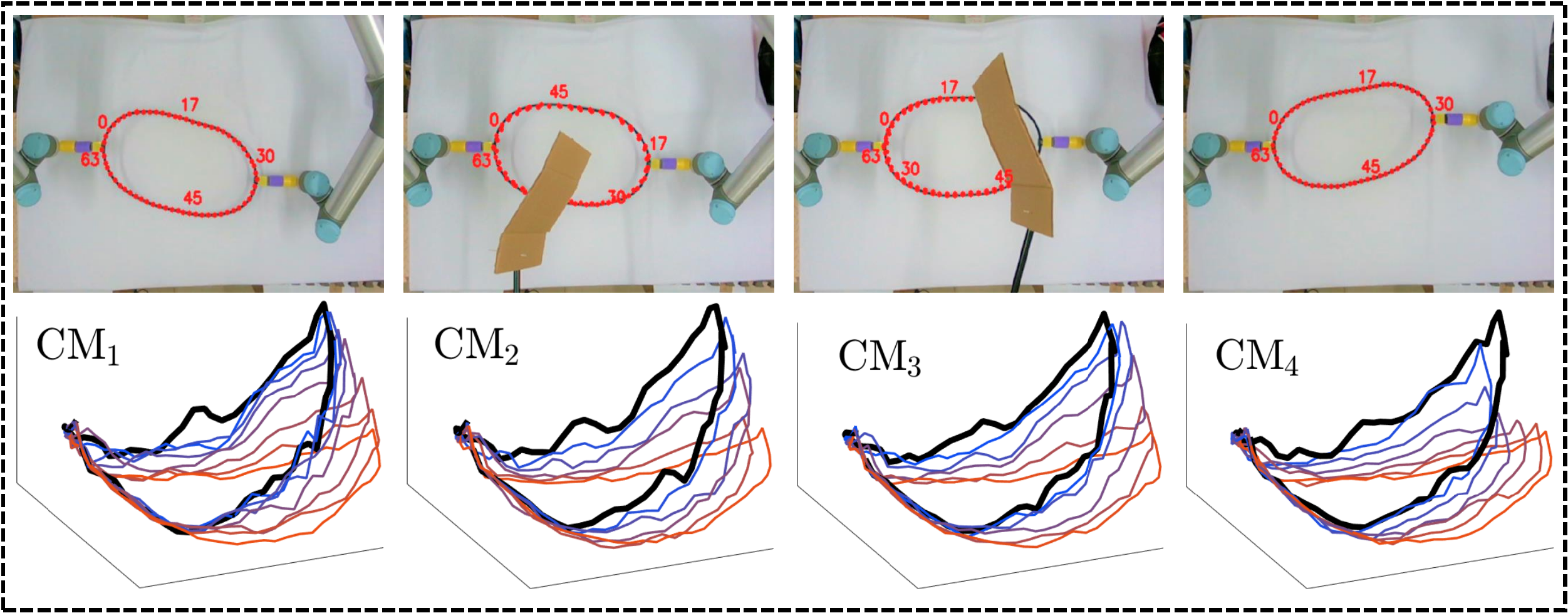}\label{fig63b}}
	
	\vspace{-0.2cm}
	\subfloat[Exp3 / Surface]
	{\includegraphics[scale=0.34]{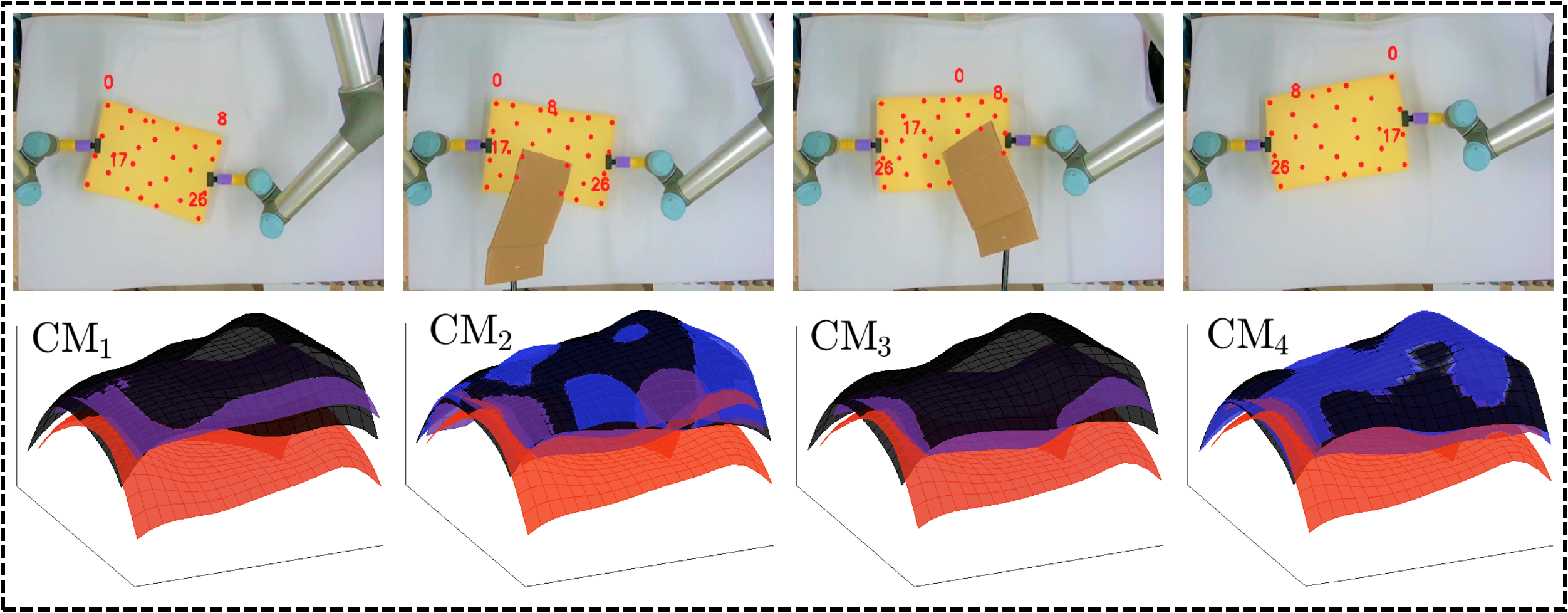}\label{fig63c}}
	\hspace{0.07cm}
	\subfloat[Exp4 / Rigid Plane]
	{\includegraphics[scale=0.34]{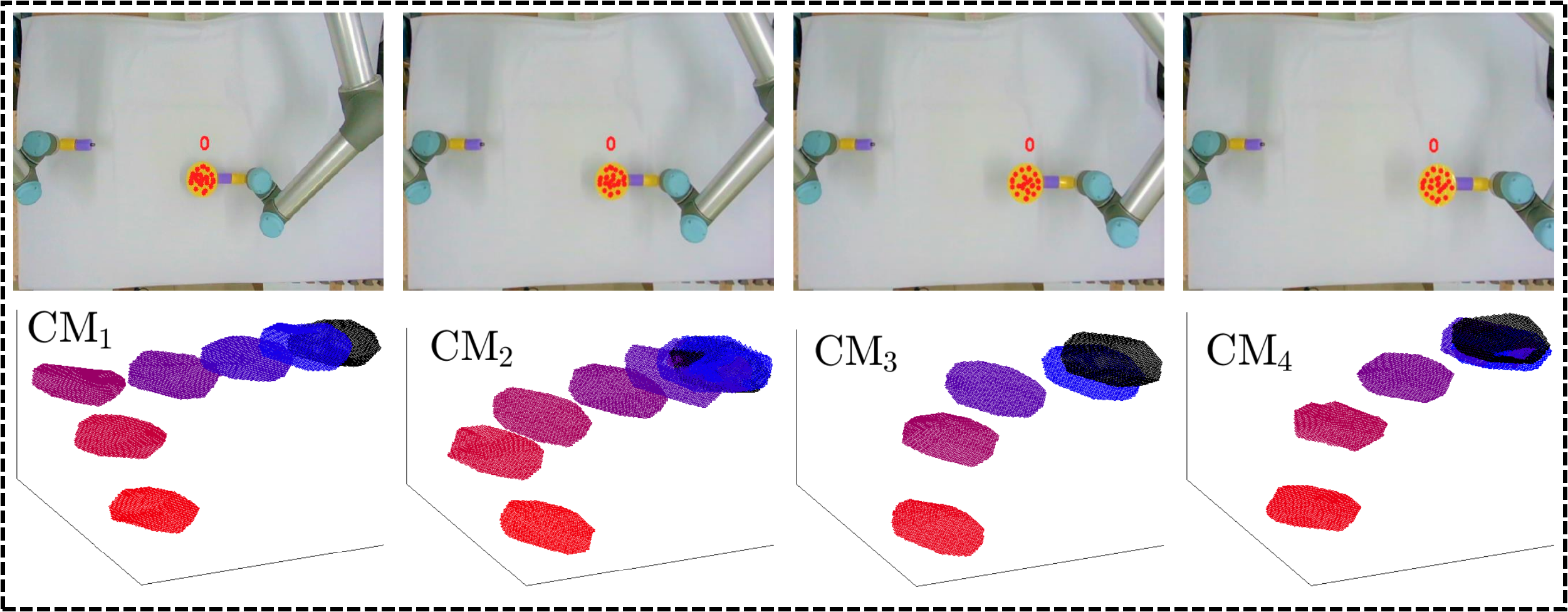}\label{fig63d}}
	
	\vspace{-0.2cm}
	\caption{
	Shape manipulation experiments, Exp1, Exp2, and Exp3 are for the deformation tasks, and Exp4 is the positioning task.
	The first row shows 2D image manipulation process (from left to right are initial, intermediate, intermediate, and desired shape), and the second row represents 3D shape manipulation process.
	The obstacle moves randomly during the process.
	The target shape is shown with black, the red represents the start shapes, and the gradient color are intermediate shapes.}
	\label{fig63}
\end{figure*}

\begin{figure*}[!htb]
\vspace{-0.5cm}
	\centering
	\subfloat[Exp1 / Centerline] 
	{\includegraphics[scale=0.465]{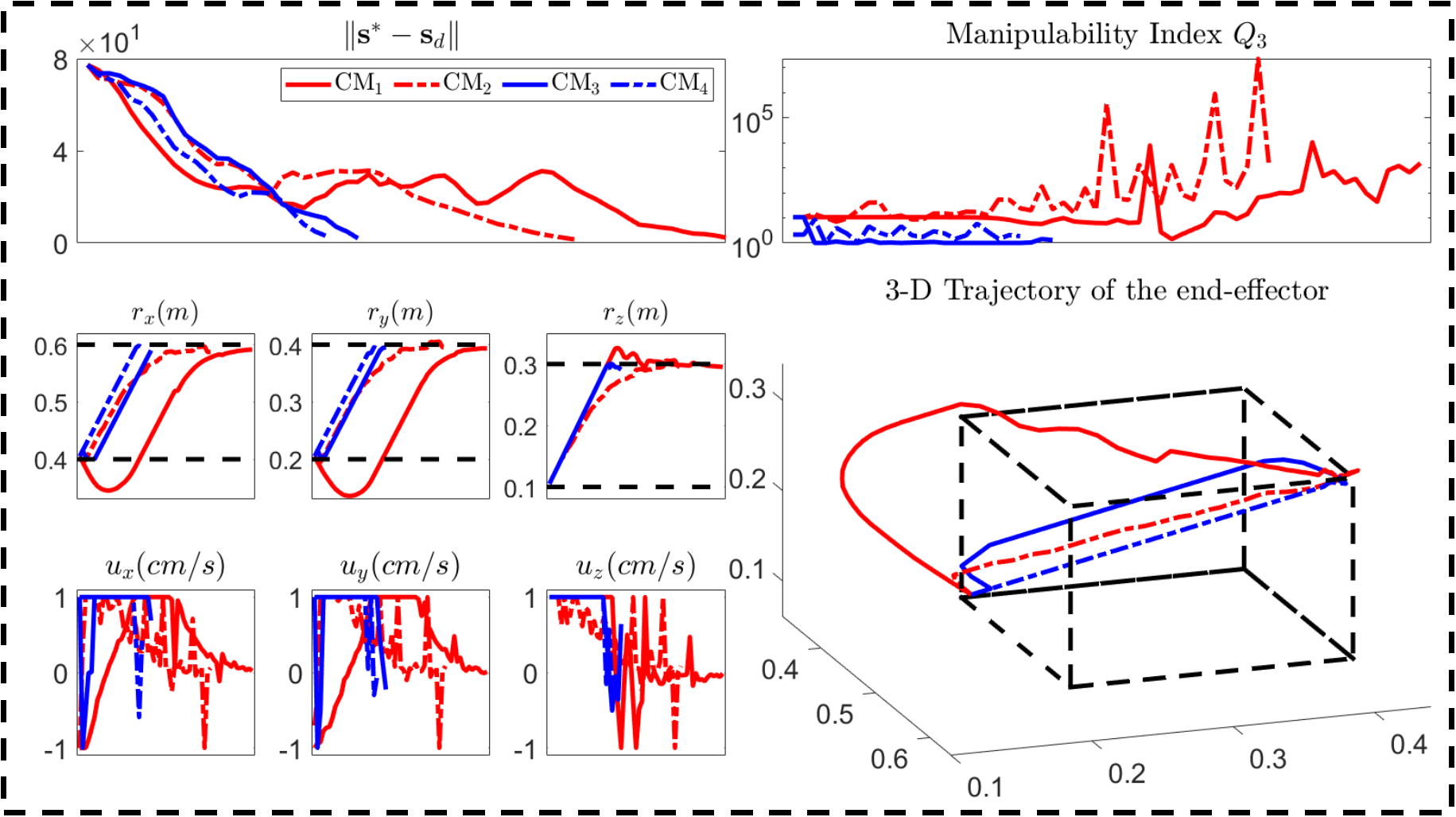}\label{fig64a}}
	\hspace{0.07cm}
	\subfloat[Exp2 / Contour]
	{\includegraphics[scale=0.465]{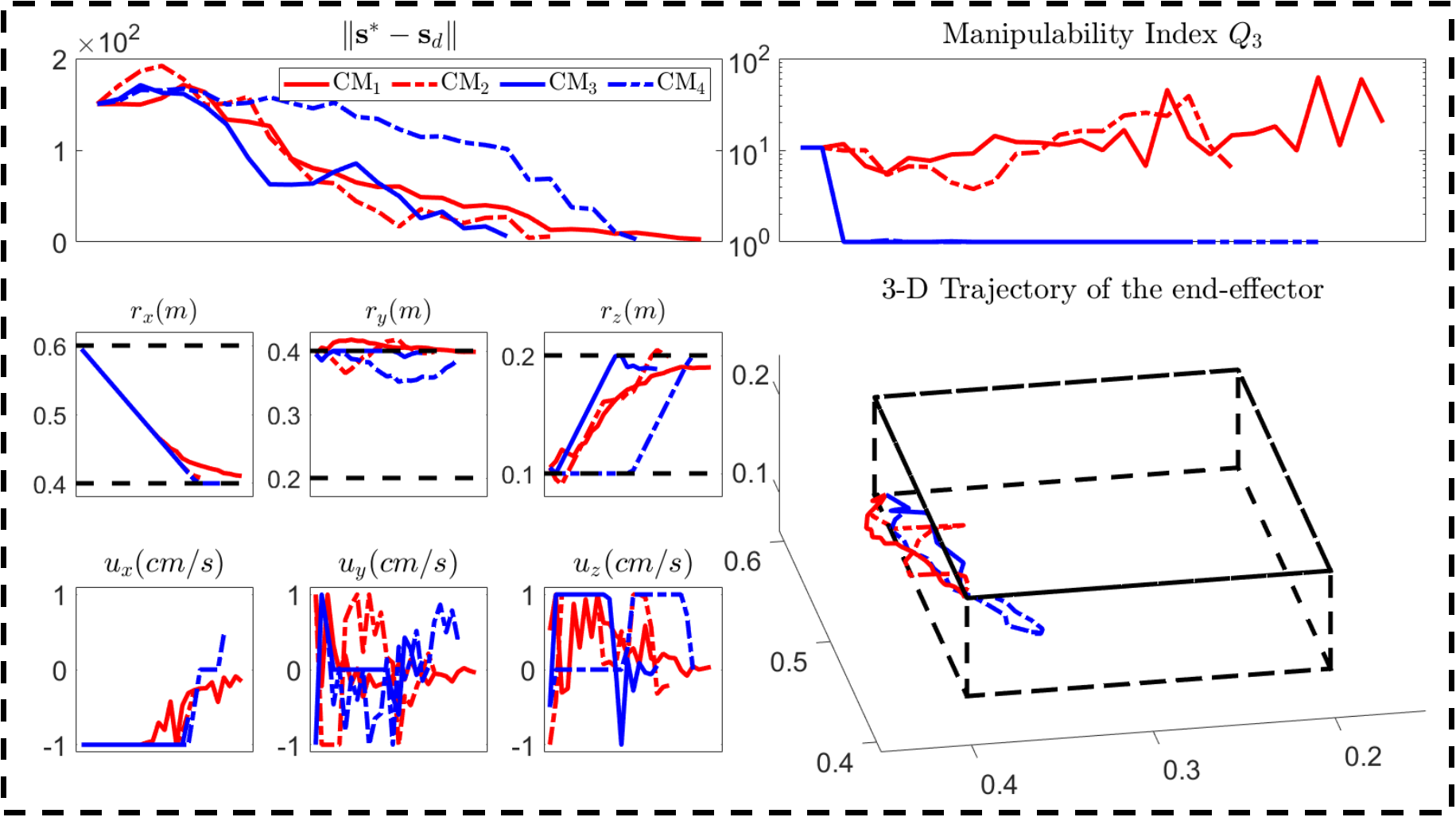}\label{fig64b}}
	
	\vspace{-0.2cm}
	\subfloat[Exp3 / Surface]
	{\includegraphics[scale=0.465]{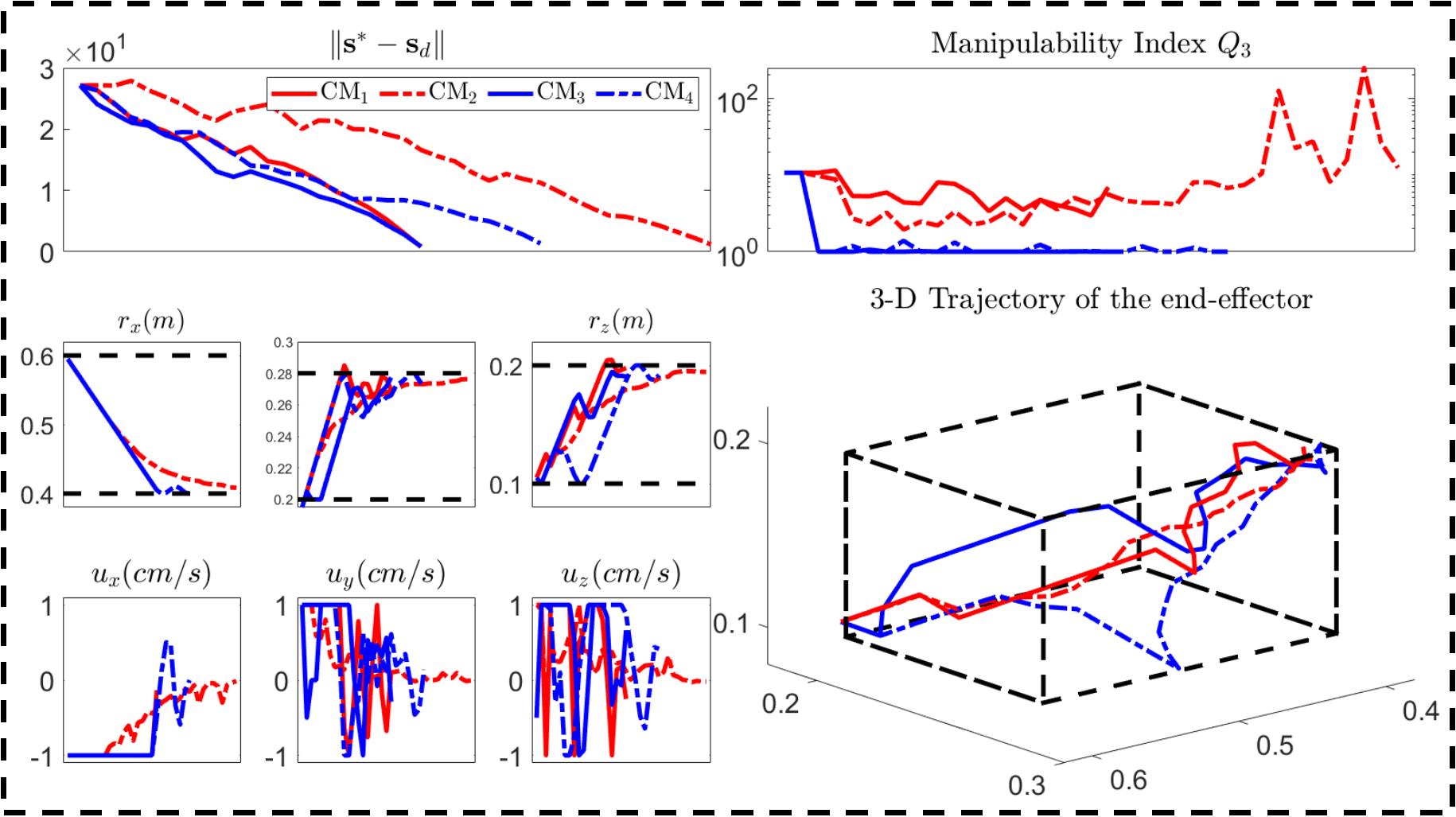}\label{fig64c}}
	\hspace{0.07cm}
	\subfloat[Exp4 / Rigid Plane]
	{\includegraphics[scale=0.465]{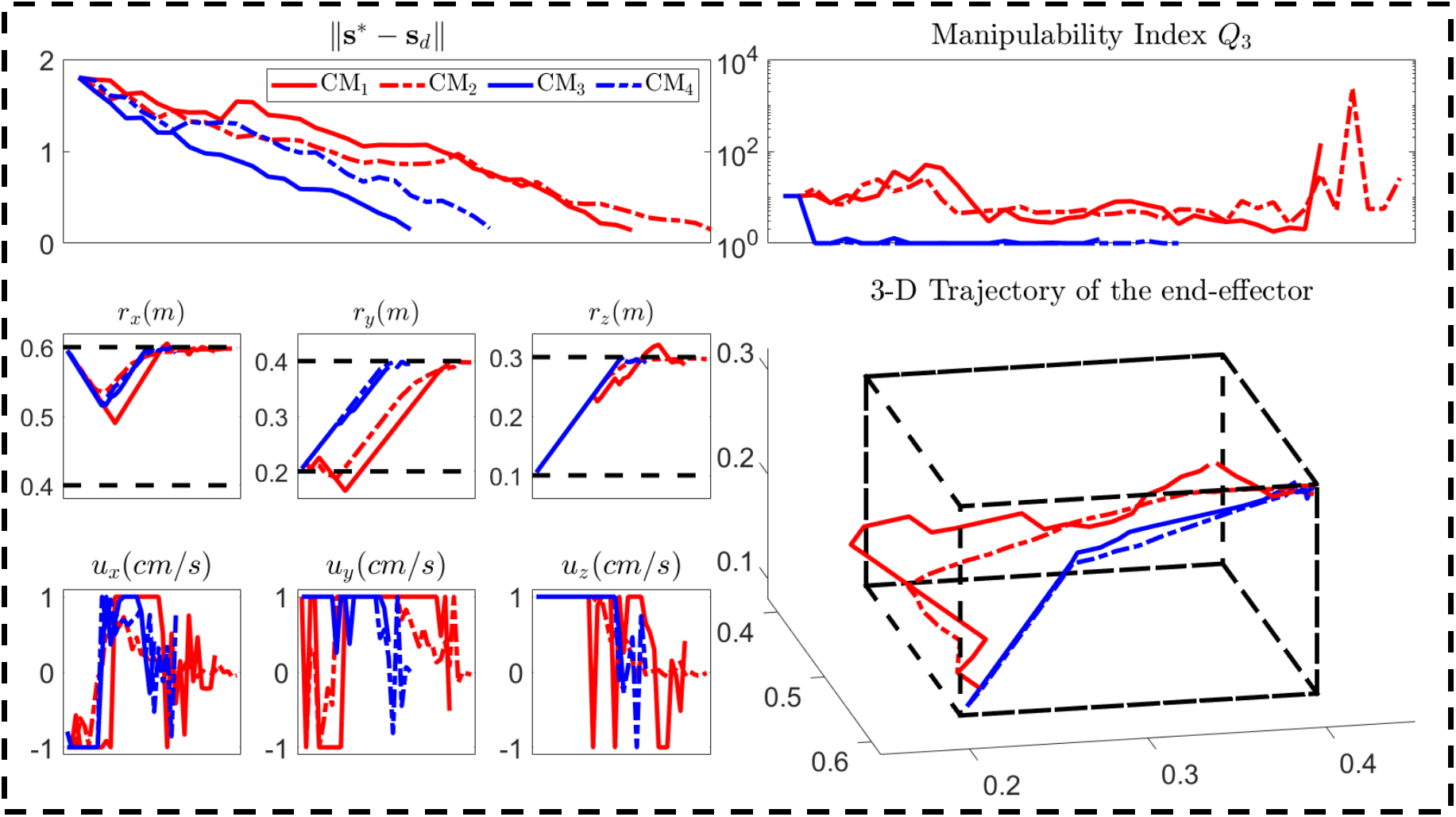}\label{fig64d}}
	
	\vspace{-0.2cm}
	\caption{
		Profiles of the manipulation error $\| \mathbf{s}^* - \mathbf{s}_k \|$, the robot pose $\mathbf{r}_k$, the velocity command $\mathbf{u}_k$, and the manipulability index $Q_3$ among four experiments (Exp1, \ldots, Exp4).
		Each experiment adopts four methods, i.e., $\rm{CM}_1, \ldots, \rm{CM}_4$.}
	\label{fig64}
	\vspace{-0.3cm}
\end{figure*}

\begin{figure}[h]
	\centering
	\includegraphics[scale=0.25]{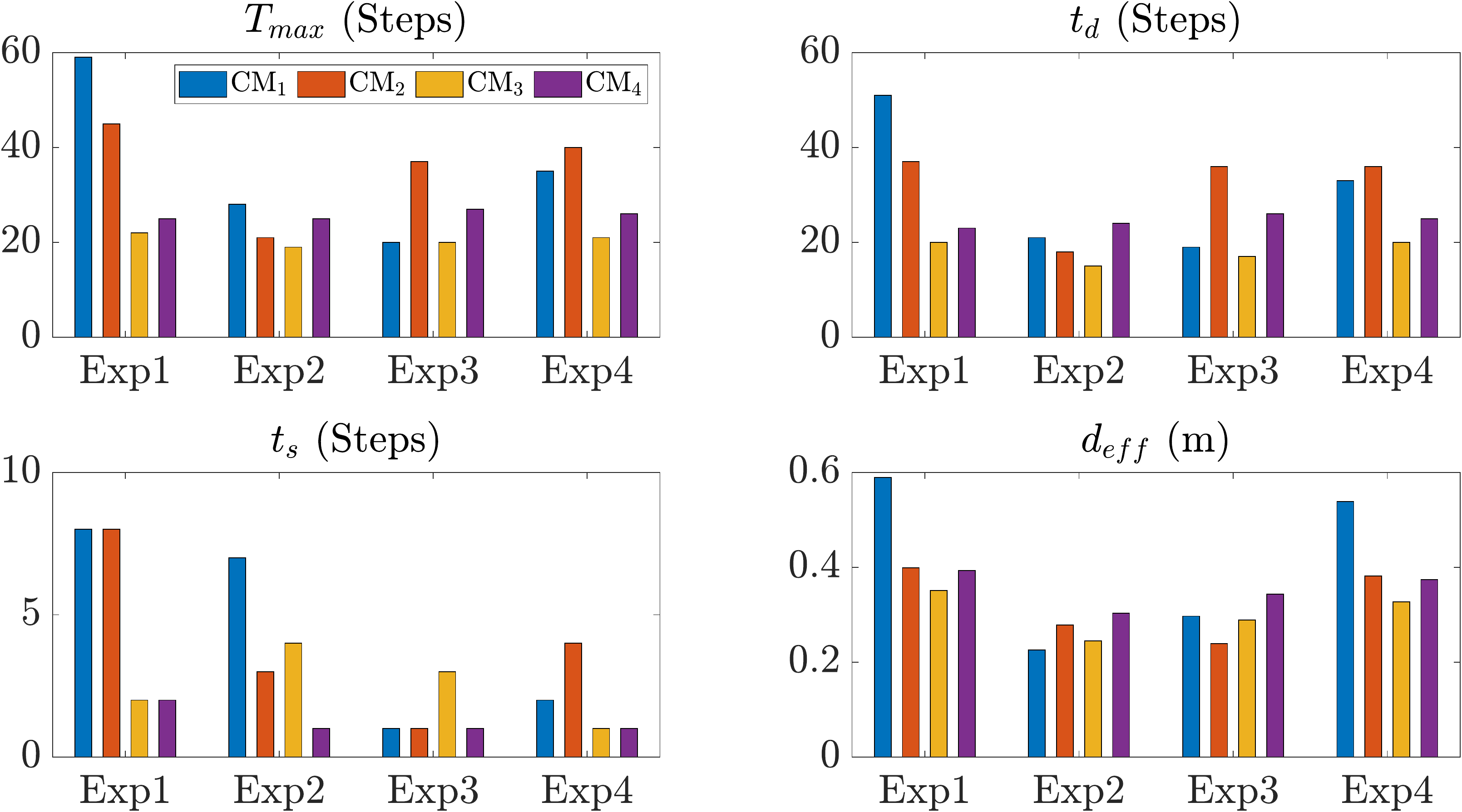}
	\vspace{-0.3cm}
	\caption{
		Performance comparison in the manipulation tasks (Exp1 to Exp4).}
	\label{fig61}
\end{figure}

\subsection{Automatic Shape Servoing Control}\label{section7f}
This section conducts four automatic manipulation experiments, labeled as Exp1, Exp2, Exp3, and Exp4, respectively.
The target contour $\bar{\mathbf{c}}^*$ is obtained from previous demonstrations of the manipulation task, which ensures its reachability.
A cardboard sheet is manually placed over the object to produce (partial) occlusions and test the robustness of our algorithm.
The estimation methods in \cite{zhu2021vision} and \cite{qi2020adaptive} are compared with the proposed receding-time model with MPC ($h=5$ and $h=15$).
We label these methods as $\rm{CM}_1, \ldots, CM_4$, and each method has been optimized to achieve a balance of stability, convergence, and responsiveness.

In addition to the feedback shape error $\|\mathbf s^* - \mathbf s_k \|$, we also compare $T_{max}$ (i.e., the number of steps from start to finish),
$t_d$ (the steps from $\| \mathbf{s}^* - \mathbf{s}_0 \|$ to 10\% of this value),
$t_s$ (the steps from 10\% $\| \mathbf{s}^* - \mathbf{s}_0 \|$ to the threshold value), and $d_{eff}$ (the total moving distance of the end-effector), \cite{qi2021contour,lagneau2020active}.
Fig. \ref{fig63} shows the shaping motions of the manipulated objects toward the desired configuration (black curve), with the cardboard blocking the view at various instances.
The results demonstrate that SPN can predict the object's shape during occlusions and feed it back to the controller to enforce the shape servo-loop; This results in the objects being gradually manipulated towards the desired configuration.
The error norm
$\| \mathbf{s}^* - \mathbf{s}_k \|$ plots in Fig. \ref{fig64} shows that $\rm{CM}_3$ provides the best control performance for the error minimization, with $\rm{CM}_4$ as the second-best, and $\rm{CM}_1$ and $\rm{CM}_2$ showing similar performance.
A higher $h$ is helpful for feature prediction, yet, since we assume that $\hat{\mathbf{J}}_k$ is constant in the window period $h$, it may lead to inaccurate predictions and even wrong manipulation of objects.
Therefore, $h$ should be chosen according to the performance requirements of the system.

From the $Q_3$ plots in Fig. \ref{fig64}, we can see RTM with $\eta=10$ provides the smallest value, which validates that RTM can enhance the manipulation feasibility and avoid shapes falling into the singular configurations.
The black dashed lines represent the workspace constraints of $r_x,r_y,r_z$, on which we can see the end-effector's trajectories in the Cartesian coordinate system; The results show that $\rm{CM}_3$ and $\rm{CM}_4$ remain within the workspace, while $\rm{CM}_1$ and $\rm{CM}_2$ may violate this constraint.
Due to the adopted input saturation in our platform, all four methods can satisfy the control saturation constraint.
Exp4 shows that our method has good universality, not only for the shape control, but also for traditional rigid object positioning.

A performance comparison of these experiments is given in Fig. \ref{fig61}.
The results illustrate that our proposed shape servoing framework achieves the best performance relative to the manipulation speed $T_{max}$, response speed $t_d$, convergence speed $t_s$, and motion distance $d_{eff}$.


\section{Conclusions}\label{section8}
In this paper, we present an occlusion-robust shape servoing framework to control shapes of elastic objects into target configurations, while considering workspace and saturation constraints.
A low-dimensional feature extractor is proposed to represent 3D shapes based on LSM and MLS.
A deep neural network is introduced to predict the object's configuration subject to occlusions, and feed it to the shape servo-controller.
A receding-time model estimator is designed to approximate the deformation Jacobian matrix with various constraints such as accuracy, smoothness, singularity.
The conducted experiments validate the proposed methodology with multiple unstructured shape servoing tasks in visually occluded situations and with unknown deformation models.

However, there are some limitations in our framework.
For example, as the support field radius $d$ is constant (i.e., it does not adjust with dynamic shapes), the computed representation lacks flexibility.
Also, the SPN needs to obtain substantial offline training data to properly work, which might pose complications in practice.
Note that our method may not accurately shape objects with negligible elastic properties (e.g., fabrics, food materials, etc).
Future work include the incorporation of shape reachability detection into the framework in order to determine the feasibility of a given shaping task beforehand.

\appendices
\ifCLASSOPTIONcaptionsoff
  \newpage
\fi

\bibliography{biblio.bib}
\bibliographystyle{IEEEtran}

\begin{IEEEbiography}
[{\includegraphics[width=1in,height=1.25in,clip,keepaspectratio]{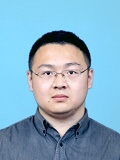}}]{Jiaming Qi}
received the M.Sc. in integrated circuit engineering from Harbin Institute of Technology,
Harbin, China, in 2018. 
In 2019, he was a visiting PhD student at The Hong Kong Polytechnic University.
He is currently pursuing the Ph.D. degree with control science and engineering, Harbin Institute of Technology, Harbin, China. 
His current research interests include data-driven control for soft object manipulation, visual servoing, robotics and control theory.
\end{IEEEbiography}

\begin{IEEEbiography}[{\includegraphics[width=1in,height=1.25in,keepaspectratio]{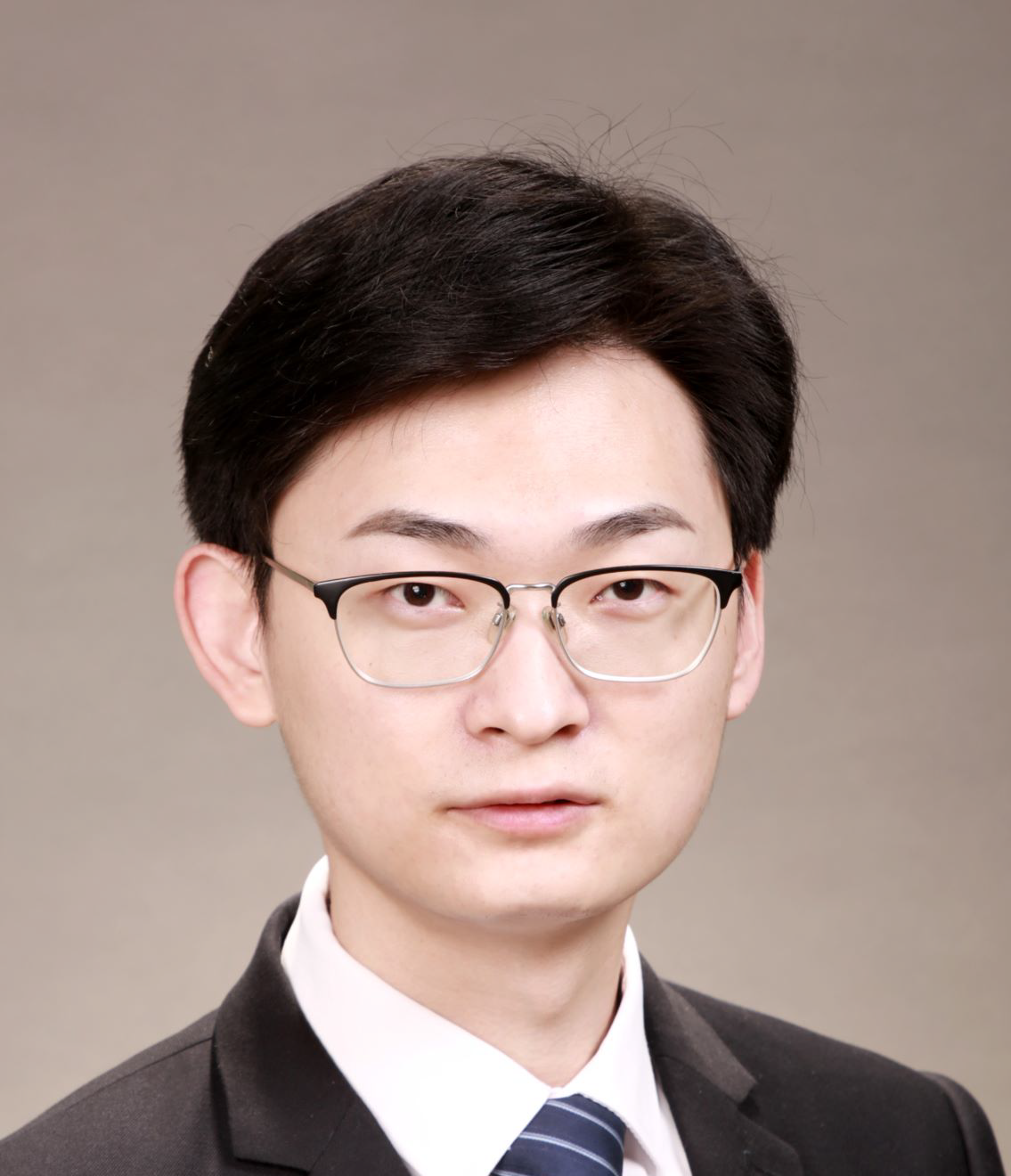}}]{Dongyu Li} (Member, IEEE) received the B.S. and Ph.D. degree in control science and engineering, Harbin Institute of Technology, China, in 2016 and 2020. He was a joint Ph.D. student with the Department of Electrical and Computer Engineering, National University of Singapore from 2017 to 2019, and  a research fellow with  the Department of Biomedical Engineering,   National University of Singapore, from 2019 to 2021. He is currently an Associate Professor with the School of Cyber Science and Technology, Beihang University, China. His research interests include networked system cooperation, adaptive systems, and robotic control. 
\end{IEEEbiography}

\begin{IEEEbiography}[{\includegraphics[width=1in,height=1.25in,clip,keepaspectratio]{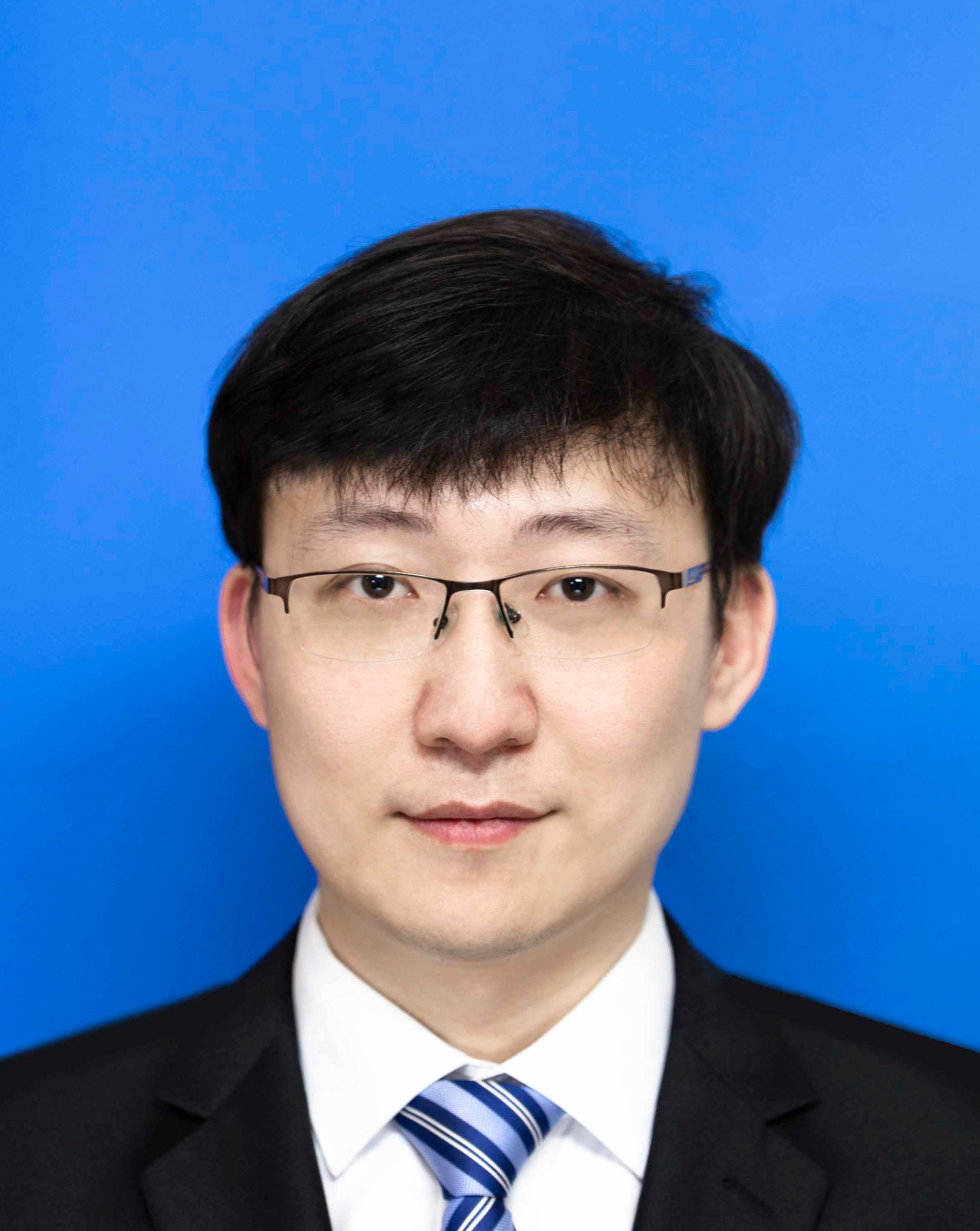}}]{Yufeng Gao}
(Student Member, IEEE) received his bachelor of engineering and bachelor of economics degrees from Wuhan University of Technology and Wuhan University, China, both in 2016. And he received the Ph.D. degree in control science and engineering, Harbin Institute of Technology, China, in 2021. He was a joint Ph.D. student with the Chair of Automatic Control Engineering, Technical University of Munich, Germany, from 2018 to 2019. 
His current research interests include spacecraft system modeling, control and optimization.
\end{IEEEbiography}

\begin{IEEEbiography}
[{\includegraphics[width=1in,height=1.25in,clip,keepaspectratio]{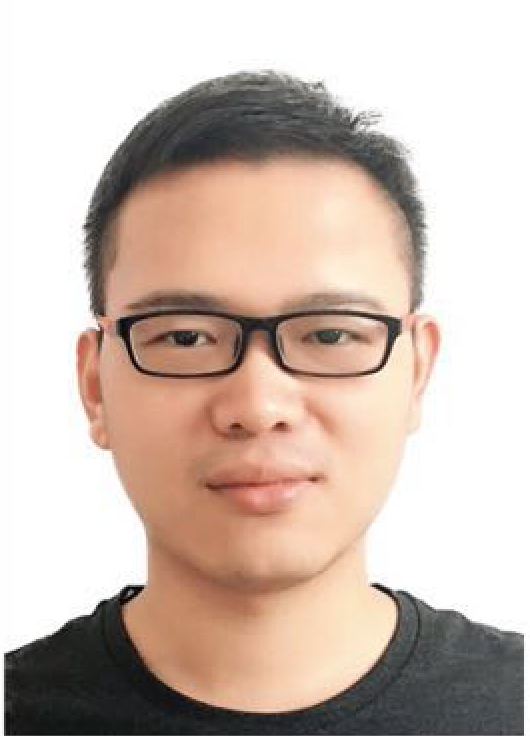}}]{Peng Zhou}
(Student Member, IEEE) was born in China. He received the M.Sc. degree in software engineering from the School of Software Engineering, Tongji University, Shanghai, China, in 2017 and is currently pursuing his the Ph.D. degree in mechanical engineering at The Hong Kong Polytechnic University, Kowloon, Hong Kong. His research interests include deformable object manipulation, motion planning and robot learning.
\end{IEEEbiography}

	\begin{IEEEbiography}
		[{\includegraphics[width=1in,height=1.25in,clip,keepaspectratio]{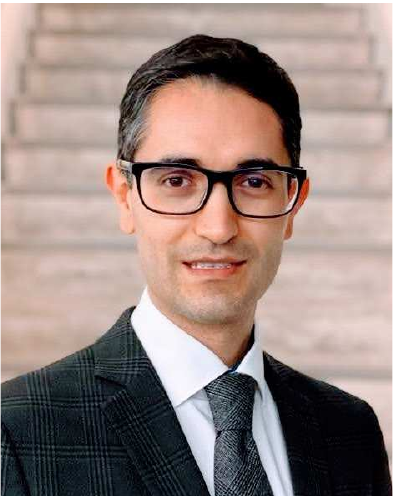}}] 	{David Navarro-Alarcon} (Senior Member, IEEE) received the Ph.D. degree in mechanical and automation engineering from The Chinese University of Hong Kong (CUHK), Shatin, Hong Kong, in 2014. 
		
		From 2014 to 2017, he was a Postdoctoral Fellow and then a Research Assistant Professor with the CUHK T Stone Robotics Institute. 
		Since 2017, he has been with The Hong Kong Polytechnic University, Hong Kong, where he is currently an Assistant Professor with the Department of Mechanical Engineering, and the Principal Investigator of the Robotics and Machine Intelligence Laboratory. 
		His current research interests include perceptual robotics and control theory.
	\end{IEEEbiography}

\end{document}